\documentclass[acmlarge]{acmart}
\usepackage{soul, listings, svg, multirow, subcaption}
\AtBeginDocument{%
  }

\setcopyright{acmlicensed}
\copyrightyear{2025}
\acmYear{2025}
\acmDOI{XXXXXXX.XXXXXXX}

\acmJournal{HEALTH}
\acmVolume{XX}
\acmNumber{X}
\acmArticle{XXX}
\acmMonth{11}

\begin{document}

\title{Dynamic Reconstruction of Ultrasound-Derived Flow Fields With Physics-Informed Neural Fields}

\author{Viraj Patel}
\email{vbp24@bath.ac.uk}
\orcid{0009-0009-1404-1093}
\affiliation{%
  \department{Department of Computer Science}
  \institution{University of Bath}
  \city{Bath}
  \country{UK}
}

\author{Lisa Kreusser}
\orcid{0000-0002-1131-1125}
\affiliation{%
  \department{Department of Mathematics}
  \institution{University of Bath}
  \city{Bath}
  \country{UK}
}
\email{lmk54@bath.ac.uk}

\author{Katharine Fraser}
\orcid{0000-0002-7828-1354}
\affiliation{%
  \department{Department of Mechanical Engineering}
  \institution{University of Bath}
  \city{Bath}
  \country{UK}}
\email{khf27@bath.ac.uk}

\renewcommand{\shortauthors}{Patel et al.}

\begin{abstract}
  Blood flow is sensitive to disease and provides insight into cardiac function, making flow field analysis valuable for diagnosis. However, while safer than radiation-based imaging and more suitable for patients with medical implants, ultrasound suffers from attenuation with depth, limiting the quality of the image. Despite advances in echocardiographic particle image velocimetry (EchoPIV), accurately measuring blood velocity remains challenging due to the technique's limitations and the complexity of blood flow dynamics. Physics-informed machine learning can enhance accuracy and robustness, particularly in scenarios where noisy or incomplete data challenge purely data-driven approaches. We present a physics-informed neural field model with multi-scale Fourier Feature encoding for estimating blood flow from sparse and noisy ultrasound data without requiring ground truth supervision. We demonstrate that this model achieves consistently low mean squared error in denoising and inpainting both synthetic and real datasets, verified against reference flow fields and ground truth flow rate measurements. While physics-informed neural fields have been widely used to reconstruct medical images, applications to medical flow reconstruction are mostly prominent in Flow MRI. In this work, we adapt methods that have proven effective in other imaging modalities to address the specific challenge of ultrasound-based flow reconstruction.
\end{abstract}




\maketitle

\section{Introduction}
Approximately 640 million people live with cardiovascular diseases globally, with 4 in 5 deaths preventable had medical treatment or lifestyle changes been implemented earlier \cite{GlobalHeartCirculatory2024}. Among various physiological indicators, blood flow dynamics have proven sensitive to structural and functional deformities caused by cardiovascular diseases. This sensitivity has enabled the development of diagnostic tools for conditions like atherosclerosis, coarctation, and arterial dissection \cite{vonreuternGradingCarotidStenosis2012,hathoutSonographicNASCETIndex2005,secchiMRImagingAortic2009}. Additionally, it has stimulated research aimed at identifying hemodynamic quantities derived from intracardiac flows that can aid in the diagnosis of heart failure \cite{zhongAssessmentCardiacDysfunction2016}. Echocardiography, a non-invasive imaging modality, plays a critical role in this context. It is suitable for patients across all age groups and remains effective even in the presence of medical implants. Moreover, it allows for real-time assessment of cardiac function and provides a means to evaluate the performance and efficiency of cardiac assist devices and prosthetics \cite{sotiropoulosFluidMechanicsHeart2016}. Unfortunately, ultrasound attenuation adds noise to the images making it difficult to confirm the presence of disease, delaying potential treatment and sometimes leading to complete misdiagnosis \cite{fatemiImprovingQualityCardiac2016}. Denoising echocardiographic flow fields may lead to faster diagnoses, facilitating earlier intervention.

Clinically, blood velocity is assessed using Doppler ultrasound. However, Doppler ultrasound is limited to a single component velocity per ultrasound transducer, meaning it can not provide complete flow fields. Instead, echocardiographic flow fields can be created by echocardiographic particle image velocimetry (EchoPIV), during which the fluid is seeded with tracer particles (usually a contrast agent containing microbubbles) to increase the scattering intensity of the blood. Cross-correlation is used to identify similar particle arrangements in consecutive frames, and the displacements of these arrangements between frames determine flow velocities \cite{forsbergClinicalApplicationsUltrasound1998}. Cross-correlation is usually performed within an interrogation window that is passed along the images \cite{nyrnesBloodSpeckleTrackingBased2020}. For high velocity flows, it is common for particles to exit the interrogation window before the next frame is captured, resulting in inaccurate measurements of blood velocity. Increasing the size of the interrogation window alleviates this, but comes at the expense of low spatial resolution of the flow field, making it difficult to identify the subtle features of the flow that indicate the presence of heart disease \cite{kimDevelopmentValidationEcho2004}.

Image denoising is a well established area in the field of computer vision, with much work focused on image-based machine learning architectures like autoencoders and convolutional neural networks (CNNs). Both of these approaches have previously been applied to flow fields with promising results \cite{duboisMachineLearningFluid2022,fukamiSuperresolutionReconstructionTurbulent2019}. The main limitation of these models is that they do not incorporate the physics of the fluid motion, so the flow field reconstruction is susceptible to unphysical hallucinations. Automatic differentiation capabilities of modern machine learning packages have facilitated the development of physics-informed neural networks (PINNs). Analytic derivatives of the neural network with respect to inputs can be calculated and used to minimise residuals of partial differential equations, encouraging the outputs of the neural network to follow governing laws of dynamics. The aforementioned image-based architectures treat the image as a discretised domain, making it difficult to obtain analytic spatial derivatives. Neural fields alleviate this issue by treating the image as a continuous spatiotemporal domain. This allows for variation in spatial and temporal resolution while upholding governing laws of the dynamics \cite{xieNeuralFieldsVisual2022}.

The quality of denoising can be attributed to two quantities: how well the reconstructed field follows the laws of physics, and how close features in the reconstructed field resemble those in the original field. Balancing the importance of these quantities remains a challenge. Adaptive learning strategies have been extensively used to help PINNs solve problems with simulated data, but these are not necessarily transferable to real-world data  \cite{gaoFailureinformedAdaptiveSampling2023,mcclennySelfadaptivePhysicsinformedNeural2023b,xiangSelfadaptiveLossBalanced2022,mcclennySelfAdaptivePhysicsInformedNeural2023}. Hard constraints, like boundary conditions, are difficult to enforce when the arterial walls are moving and the nature of ultrasound images, which are composed of speckle patterns, causing walls to look inconsistent.  In what follows, we will compare various neural field implementations for reconstructing simulated and real ultrasound velocity fields of arterial flow with varying levels of noise. The primary aim of this paper is to adapt recent advances in neural fields for visual computing - techniques that have largely been demonstrated on idealised or benchmark datasets - to the domain of real ultrasound imaging \cite{xieNeuralFieldsVisual2022}. In doing so, this work seeks to bridge the gap between highly controlled simulated data and the complexities of real-world medical imaging. Given the lack of ground truth ultrasound data, a multi-modal verification pipeline was used to determine which architecture yielded the most effective reconstructions in the context of cardiovascular diagnosis and evaluating surgical procedures and device implants.

\section{Related Work}

\subsection{Echocardiographic Particle Image Velocimetry (EchoPIV)}
Particle image velocimetry (PIV) is a flow visualisation technique based on the displacement of small groups of particles between two points in time \cite{harmandReviewFluidFlow2013}. While there are many variations of the technique, usually the fluid is seeded with small tracer particles, illuminated with a laser light sheet, and imaged with a high speed camera. flow velocities are determined by the displacement of the particles between frames and a 2D vector map of the flow velocities, called a flow field, is produced \cite{crapperFlowFieldVisualization2000}. When this technique is applied to ultrasound images, it is called echocardiographic particle image velocimetry (EchoPIV), and contrast agents consisting of microbubbles ($<10\mu$m in diameter) are used to increase the scattering intensity of the blood flow \cite{forsbergClinicalApplicationsUltrasound1998}. It is assumed that the microbubbles move at the same velocity as the fluid (flow homogeneity) \cite{walkerVitroPoststenoticFlow2014}.

\begin{figure}
    \centering
    \includegraphics[width=\linewidth]{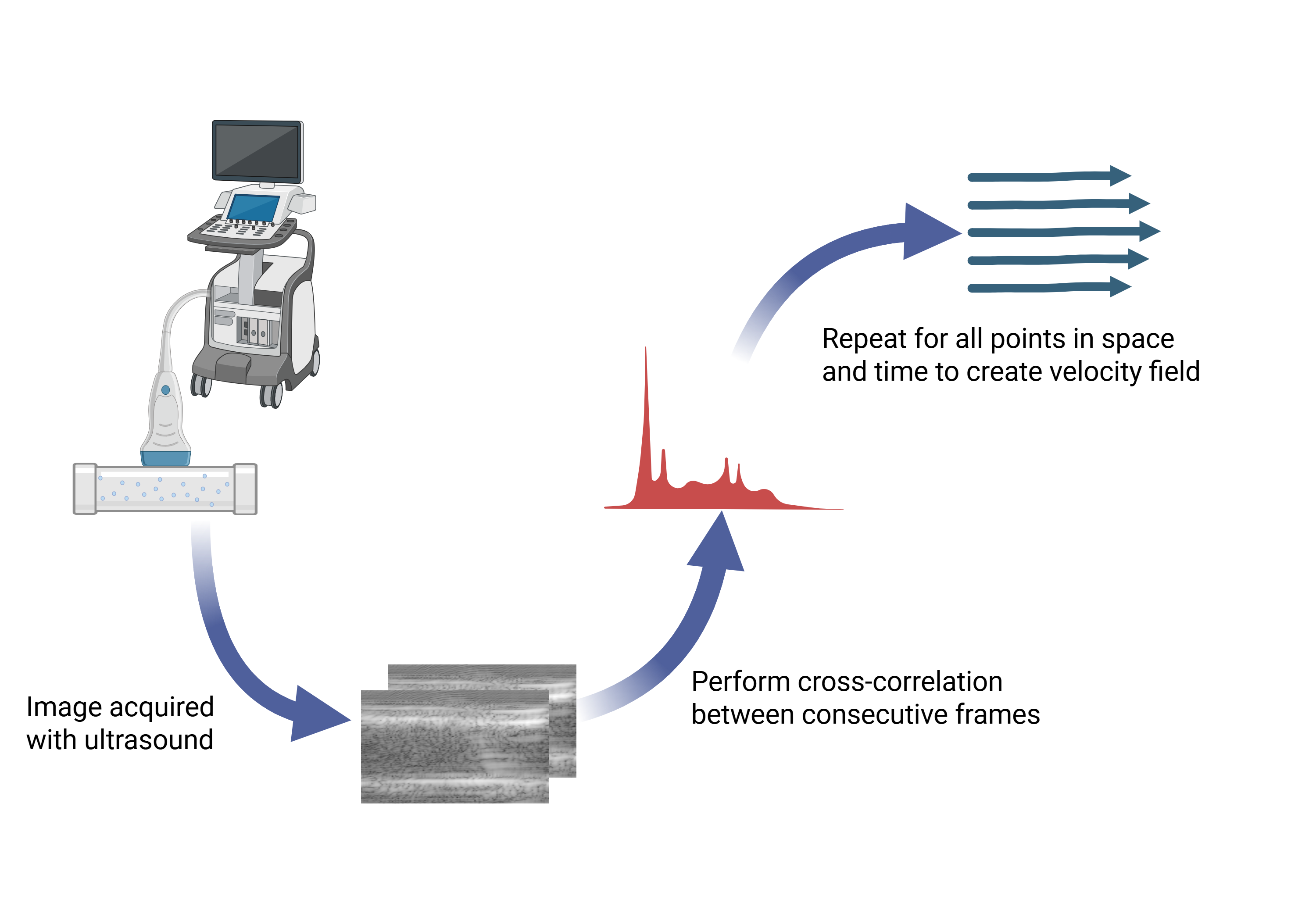}
    \caption{Schematic of EchoPIV process \cite{patelCreatedBioRender2026}. The transducer captures an ultrasound image of the blood vessel at different time steps. Interrogation windows are overlaid in the same location of consecutive frames, the cross-correlation is evaluated to infer displacement. These displacements are used to construct a flow field.}
    \label{fig:uiv1}
    \Description[Schematic of EchoPIV process]{At regular time intervals, images are acquired with an ultrasound transducer, which are then cross-correlated in pairs to infer tracer particle displacement.}
\end{figure}

As shown in Figure \ref{fig:uiv1}, raw ultrasound video recordings are obtained and broken down by frames. An interrogation window is overlaid on a specific area of one frame and the corresponding area in the next frame. The algorithm performs cross-correlation between these interrogation windows to identify common patterns and measure the displacement of particles between frames. The interrogation window is passed along the image, and the process repeats for all frames \cite{kimDevelopmentValidationEcho2004}. This allows for the calculation of a velocity field based on the observed displacements and the frame rate of the video. The frame rate, the size of the interrogation window, and the particle tracking have all been investigated and modified during the development of the algorithm.

High frame rates are essential for improved resolution in plane wave ultrasound imaging and, when applied to EchoPIV, enhance temporal resolution and expand the measurable velocity range \cite{forsbergClinicalApplicationsUltrasound1998}. At high velocities, particles may leave the interrogation window before the next frame is captured, reducing inter-frame correlation and impairing particle tracking; this underestimation effect has been observed in left ventricle phantom flow reconstructions \cite{hongCharacterizationQuantificationVortex2008}. Beyond increasing frame rate, adjusting the interrogation window size can mitigate this issue. Windows that are too small contain too few particles for reliable velocity estimates, whereas large windows introduce errors from local flow gradients, creating a trade-off between spatial resolution and measurable velocity range \cite{kimDevelopmentValidationEcho2004}. An optimal window size can be determined by iteratively reducing its dimensions until the correlation drops below a threshold \cite{hartSuperresolutionPIVRecursive1999}. This method has been improved by offsetting the interrogation window in the second frame to account for expected particle displacement, though it has yet to be applied to real particle images \cite{westerweelEffectDiscreteWindow1997}.

In practice, EchoPIV has been verified for measuring steady and pulsating velocity profiles in arteries \cite{kimNoninvasiveMeasurementSteady2004}. Fraser et al. developed a method to increase the dynamic range of EchoPIV by interleaving the images and applied this to pulsatile flow for the first time \cite{fraserUltrasoundImagingVelocimetry2017}. Their method used a multi-pass PIV algorithm starting with a large interrogation window of $32\times64$ pixels, slowly decreasing with every iteration to $4\times8$ pixels with a 50\% overlap in the final iteration. This iteratively decreasing interrogation window allows the PIV algorithm to capture flow features at a finer resolution than previous implementations \cite{hartSuperresolutionPIVRecursive1999}. In the first pass of the algorithm, a single PIV iteration was applied to the region of interest to estimate the velocity magnitude over time for all acquired frames. This estimated signal was used to align each cycle of the pulsatile flow, with cross correlation used to calculate the phase shift for each image pair. The phase shift was then used to align the raw data into cycles before the second multi-iteration pass of PIV was applied. The alignment of the pulsatile flow cycles allowed for easier correlation averaging, where the cross correlation peaks from the PIV were averaged to smooth effects of random noise in the velocity estimation. A velocity correction scheme was applied to account for the sweeping beam nature of conventional, focused ultrasound \cite{zhouUltrasoundImagingVelocimetry2013a}. Since transducer elements acquire columns sequentially, full-frame construction takes time, causing tracer particles to shift between first and last column acquisition. The system timestamps the frame at acquisition start, so reported particle positions and velocities actually correspond to later times, not the true frame start. To correct this, tracer velocities from the PIV algorithm and accelerations (via central differences) were combined with the beam sweeping velocity to estimate true flow velocities at frame onset. Fraser et al. showed that when sweeping velocity matched flow velocity, measured velocity diverged to infinity; this effect was absent for negative sweeping velocities, so they set the sweep opposite to the flow.

Although these adaptations to the EchoPIV methodology have led to some improvements, the fundamental limitations of ultrasound, such as dropout and signal occlusion, remain unavoidable, particularly at depth which is of great concern for cardiovascular imaging. Despite the drawbacks of EchoPIV, it has proven to be the best particle tracking technique for ultrasound imaging of fluid flow. However, the application of machine learning to EchoPIV has not yet been explored, so its potential remains uncertain. In this work, we will investigate the use machine learning to reconstruct flow fields in post-processing. The data set from Fraser et al. covers a range of Womersley flows under different imaging conditions. The scope of the experiments make this data suitable for use here to investigate the potential for physics-informed neural fields in denoising cardiovascular flow fields.

\subsection{Physics-Informed Machine Learning}
Many optimization problems require differentiating an objective function with respect to its parameters. Traditional approaches such as manual, numeric, or symbolic differentiation are often impractical for neural networks, which lack explicit functional forms and scale poorly for higher-order derivatives. Modern machine learning frameworks address this through automatic differentiation, which constructs computational graphs of arithmetic and transcendental operations to enable exact derivative computation \cite{baydinAutomaticDifferentiationMachine}. This technique is central to physics-informed learning, as it allows accurate evaluation of differential equation terms. Physics-Informed Neural Networks (PINNs) have been widely applied across physical systems, with the Navier–Stokes FlowNet (NSFNet) by Jin et al. representing a key development in fluid dynamics \cite{jinNSFnetsNavierStokesFlow2021, leiteritzHowAvoidTrivial2021}. Unlike conventional networks trained solely to minimize data fidelity losses, PINNs include an additional physics loss enforcing the residuals of governing equations. Through automatic differentiation, NSFNets compute exact spatial and temporal derivatives from space–time inputs to predict velocity and pressure fields while incorporating boundary condition terms to avoid trivial solutions. The total loss function generally takes the form
\begin{equation}
\mathcal{L} = \lambda_\text{data}\mathcal{L}_\text{data}+\lambda_\text{bc}\mathcal{L}_\text{bc}+\lambda_\text{phys}\mathcal{L}_\text{phys}
\end{equation}
where $\mathcal{L}_\text{data}$, $\mathcal{L}_\text{bc}$, and $\mathcal{L}_\text{phys}$ are the data fidelity, boundary condition, and physics loss terms, respectively. Cuomo et al. discuss a range of applications of physics-informed machine learning from recurrent neural networks being used for numerical integration, to Bayesian machine learning used to model stochastic processes \cite{cuomoScientificMachineLearning2022}.

\subsection{Flow Field Reconstruction}
While significant efforts have been made to enhance velocimetry and particle tracking algorithms in fluid dynamics through machine learning, such approaches are often ineffective when limitations arise from imaging quality rather than algorithmic performance; in these cases, machine learning must instead be employed to post-process and reconstruct the flow fields from the degraded measurements \cite{zhenMachineLearningApproach2024}. Prior work has addressed this challenge using inverse problems and variational approaches, but these models are constrained by the need for prior knowledge of the flow, which is not always available in clinical settings, particularly for patients with cardiovascular diseases. This drawback could be seen in the work of Arigovindan et al., who applied total variation denoising to arterial flows obtained by pulsed wave Doppler ultrasound and used predicted motion to incorporate physics into the model \cite{arigovindanFullMotionFlow2006}. Meyers et al. attempted to work around this problem by informing the model with the Navier-Stokes equations to increase the generalisability, but their model was found to have oversimplified the motion \cite{meyersColourDopplerEchocardiographyFlow2020}. Kontogiannis et al. proposed a physics-informed inverse problem formulation to denoise phase-contrast MRI (PC-MRI) flow fields but their approach was only applicable to flow fields obtained by PC-MRI \cite{kontogiannisPhysicsinformedCompressedSensing2022, kontogiannisBayesianInverseNavierStokes2024}. In the scope of ultrasound imaging, Jang et al. formulated an inverse problem to obtain a vector field from colour Doppler scans of a left ventricle phantom, though the work relies on ultrafast frame rates common with experimental ultrasound systems but that may not yet be applicable to clinical machines \cite{jangReconstructionMethodBlood2015}.

Deep learning approaches have been proposed due to the complexity of biomedical flows and the ability to generalise over vast datasets. Inspired by various works using autoencoders to denoise images, Dubois et al. found that, when applied to flow fields, linear autoencoders were susceptible to overfitting to noise and variational autoencoders alleviated this issue at the cost of lower fidelity \cite{duboisMachineLearningFluid2022}. Moreover, their architecture was limited by the resolution of the flow field. Fukami et al. successfully used CNNs to perform super-resolution on turbulent flow fields and ensemble methods have extended their work to reconstruct and upscale flow MRI data \cite{fukamiSuperresolutionReconstructionTurbulent2019,ericssonGeneralizedSuperResolution4D2024}. All of these models used image-based architectures that rely on discretised space, making it difficult to enforce physical laws accurately as automatic differentiation is usually not possible and numerical differentiation is prone to large errors. Statistical approaches to denoising, like a diffusion model proposed by Shu et al., can use automatic differentiation to enforce physical laws but have been found to struggle with enforcing hard constraints like boundary conditions \cite{shuPhysicsinformedDiffusionModel2023}. They also discuss how the fidelity of the model is limited by the sampling procedure and volume of training data, which pose a significant challenge in experimental and clinical contexts where data is often limited and sparse.

In parallel, several studies have explored the use of neural networks to denoise and reconstruct flow fields while enforcing physical laws and using continuous space. Akbari et al. reconstructed noisy CFD data with neural networks that took in space and time coordinates and predicted velocity vectors - an example of a neural field \cite{akbariFlowBasedFeatures2021}. To incorporate physics, they extracted derivative-based features, like the divergence and curl, from the noisy input flow fields and assessed their presence in the reconstructed outputs, using these physically meaningful quantities to encourage higher fidelity and promote consistency with underlying flow physics. Their approach relied on large volumes of training data for the same flow conditions and, for very noisy data, the derivative-based features may be corrupted. Sautory et al. used an autoencoder to extract features that were fed into a PINN for reconstructing CFD simulations of vascular flow with additive Gaussian noise \cite{sautoryUnsupervisedDenoisingSuperResolution2024}. Their approach - an example of a conditional neural field - was capable of super resolution, but the additive Gaussian noise is not representative of noise obtained in medical imaging, like echocardiography. Moreover, autoencoders are not robust to noise in the latent space, as discovered by Dubois et al. \cite{duboisMachineLearningFluid2022}. Neural fields are a prospective approach to flow field reconstruction as they offer adjustable spatiotemporal resolution and are able to use automatic differentiation to incorporate physics. We wish to extend this beyond the realm of simulations and tackle the limitations associated with real-world data.

\subsection{Neural Fields}
It has long been known that a neural network, $\mathcal{N}$, can be treated as a universal function approximator, mapping between two different finite-dimensional spaces $\mathcal{N}: \mathbb{R}^n \to \mathbb{R}^m$, whether it is between different dimensions or different resolutions \cite{hornikMultilayerFeedforwardNetworks1989}. When this mapping describes a quantity varying over space and time, it is called a neural field $\mathcal{N}_f: (\vec{x},t) \to \mathbb{R}^m$. So, spatially varying data-types like images and videos can be represented by neural fields. In these applications, the data is no longer stored in its typical grid-based structure, but instead is stored in a tabular fashion with the pixel location and pixel values (usually an RGB vector) as columns and each entry is a different pixel. For videos, the frame or timestamp is also recorded. This format allows the neural field to learn a mapping from the pixel domain to the pixel value domain, treating the image or video as a function of spatiotemporal coordinates. The neural field can then be evaluated at coordinates that were not in the training set, allowing one to vary the resolution of the image \cite{xieNeuralFieldsVisual2022}.

When coordinates are directly used as inputs, neural networks have been shown to exhibit spectral bias to low-frequency signals \cite{gromniakNeuralFieldConditioning2023}. Often, high-dimensional data lies on a lower dimensional manifold due to physical and semantic constraints. When this is the case, the shape of the manifold affects the learnability of high frequencies in the target function as these low dimensional manifolds typically have low frequencies, which neural networks learn faster. Focusing on low frequencies allows the neural networks to generalise better, but comes at the cost of not capturing finer details well. Embedding the coordinates with a function that has high frequency components increases the network's expressibility of high frequency data \cite{rahamanSpectralBiasNeural2019}. Tancik et al. proposed an embedding technique called Random Fourier Features (RFF) where the inputs of the neural network $\vec{v} = (t,\vec{x}) \in \mathbb{R}^n$ are transformed by an embedding function $\gamma: \mathbb{R}^n \to \mathbb{R}^{2d}$, $d > n$, defined by
\begin{equation}
\mathcal{F}(\vec{v}) = [\cos(2\pi \vec{b_1}^\top\vec{v}), \sin(2\pi \vec{b_1}^\top\vec{v}), \cdots, \cos(2\pi \vec{b_d}^\top\vec{v}), \sin(2\pi \vec{b_d}^\top\vec{v})]^\top = [\cos(2\pi B \vec{v}), \sin(2\pi B \vec{v})]^\top
\label{eq:rff1}
\end{equation}
where the matrix $B = (\vec{b_1}, \vec{b_2}, \cdots, \vec{b_d}) \in \mathbb{R}^{n\times d}$ is sampled from a normal distribution with mean $0$ and standard deviation $\sigma$, and the $\sin$ and $\cos$ functions are applied element-wise \cite{tancikFourierFeaturesLet2020}. While both RFF and PhaseDNN, proposed by Cai et al. \cite{caiPhaseShiftDeep2020}, aim to capture high-frequency components in the data, RFF avoids the need for evenly spaced meshes and densely populated data, which PhaseDNN relies on through its pre-determined phase shift approach. It has been demonstrated with simulated data that neural fields, particularly with physics constraints, are effective models for flow field reconstruction in various domains, ranging from aerospace flows \cite{akbariFlowBasedFeatures2021} to biomedical flows \cite{sautoryUnsupervisedDenoisingSuperResolution2024}. When extending to real medical scans, however, it has been shown that Fourier encodings and further constraints in the loss function may improve transferability \cite{garziaNeuralFieldsContinuous2024}.

\subsection{Contributions}
The contributions of this work can be summarised as follows:
\begin{itemize}
    \item Constraining the neural field with the Navier-Stokes, and applying this to simulated and real datasets.
    \item Comparing different types of Fourier Feature encodings, and identifying architectures that performed well on both simulated and real datasets.
    \item Inpainting occluded flow fields with a branched neural field model that combines different types of Fourier Feature mapping strategies.
    \item Presenting a multi-modal verification pipeline to handle the lack of ground truth flow field data.
\end{itemize}

\section{Datasets}

\subsection{Pulsatile Flow and Synthetic Data Generation}
Our experiments were conducted on pulsatile flow through a long, straight tube, chosen for its well-established analytical solution and its relevance as a canonical model for arterial flow. Womersley found the analytic solution of the flow field
\begin{equation}
u(y,t) = \Re\left\{ \sum_{n=0}^{N} \frac{iP_n'}{\rho nf} \left[1-\frac{J_0(\alpha yn^{1/2}i^{3/2})}{J_0(\alpha n^{1/2}i^{3/2})}\right] e^{inf t}\right\}
\label{eq:wom1}
\end{equation}
where $u$ is the velocity in the $x$-component, $\rho$ is the density of the fluid, $y = \frac{r}{R}$ is the ratio between the radial coordinate $r$ and the radius of the tube $R$, $J_0(\cdot)$ is the Bessel function of first kind order 0, $P_n'$ is the pressure gradient magnitude of the $n$th mode, $f$ is the angular frequency of the first harmonic of the pressure gradient, and $\alpha=R\left(\frac{f\rho}{\mu}\right)^{1/2}$ is the dimensionless Womersley number for viscosity $\mu$ \cite{womersleyMethodCalculationVelocity1955}.

The neural field models were tested on synthetic data before real ultrasound data was used. A pipeline was created that first generated black and white particle images based on Womersley flow, and then PIV was performed on these images to create a flow field \cite{probstSynpivimageSynpivimage2024,liberzonOpenPIVOpenpivpython2021}. The two main sources for ultrasound attenuation are the absorption and scattering of the ultrasound waves when interacting with a medium. Although the affect of absorption is not well understood for biological tissues, scattering is known to produce speckled patterns for soft tissue and moving solid tissue \cite{laugierIntroductionPhysicsUltrasound2011,wellsUltrasoundImaging2006}. Although the statistics of the speckle noise is intrinsic to the material that the wave is scattering off of, the spatial distribution of the speckles in the image appear random due to the attenuation of these scattered signals, decreasing the signal-to-noise ratio \cite{michailovichDespecklingMedicalUltrasound2006}. To simulate this effect, noisy synthetic data was generated by adding different levels of speckle noise to the images before performing PIV. Building on previous work that improved the accuracy of EchoPIV in \cite{fraserUltrasoundImagingVelocimetry2017}, similar conditions were simulated. The radius of the pipe was modelled as $2.5$mm, the density of blood was $1060$kg m$^{-3}$, and the viscosity of blood was $3\times10^{-3}$Pa s. The Womersley number of the flow is sensitive to the density, viscosity, and radius of the blood vessel, so suggestions from literature were used to inform the choice of $\alpha = 2.77$, and $f$ was calculated from this choice \cite{vergaraWomersleyNumberbasedEstimation2010}. As EchoPIV does not measure pressure fields, the velocity-vorticity formulation of the Navier-Stokes equations proposed by \cite{trujilloPenaltyMethodVorticity1999} was used, defined by
\begin{equation}
    \frac{\partial\vec{\omega}}{\partial t} + \nabla \times (\vec{\omega} \times \vec{u})=\frac{1}{Re} \nabla^2 \vec{\omega}
\label{eq:ns1}
\end{equation}
where $\vec{u}$ is the velocity, $\vec{\omega} = \nabla \times \vec{u}$ is the vorticity, and $Re$ is the Reynolds number. This formulation has been used in NSFNets, and a Reynolds number of $500$ was chosen as is typical in carotid arterial flow \cite{ghalichiLowReynoldsNumber1998}.

\subsection{Real Ultrasound Data}
Fraser et al. scanned a 5mm diameter latex tube suspended in a tank of water, pumped with water or a water-glycerol mixture, with an acoustic absorber placed at the bottom of the tank \cite{fraserUltrasoundImagingVelocimetry2017}. The working fluid, pulsing at 60 bpm, was seeded with a microbubble-based contrast agent that produced bubbles in the range of 1-7$\mu$m in diameter. In this work, data from phantoms A (water-glycerol mixture, $Re = 500$) and E (water only, $Re = 2695$) were used as they were of similar frequency to a human heart at 60bpm. In \cite{poelmaEnhancingDynamicRange2013}, Poelma et al. discussed how a conventional Doppler ultrasound transducer with $J$ elements will have a time difference between consecutive frames of $\Delta t = J\tau$, where $\tau$ is the time taken to produce each line - usually determined by the depth and speed of sound in the medium. To increase the temporal resolution, they proposed to start recording the next frame after $m$ lines of the prior frame have been recorded, reducing the time difference to $\Delta t = m\tau$ where $m \leq J$. This interleaved technique was used by Fraser et al. to show that lower $m$ resulted in better agreement with the flow rate measured by a transit time flow meter. There is, however, a limit on how small $m$ can be (dependent on the capabilities of the ultrasound scanner), resulting in occluded regions in the flow field where the flow has changed before the full scan has been collected. Although the source of occlusions in the flow field data used in this work is somewhat outdated, since plane-wave ultrasound imaging has largely replaced earlier acquisition methods, the broader problem of incomplete or occluded flow fields remains highly relevant. Occlusions can arise from several factors, including non-uniform contrast agent distribution, acoustic shadowing, non-planar surfaces, or the limited field of view imposed by the finite transducer size. Here, the outdated source of occlusions is used only as a representative example. Importantly, the methods we propose for reconstructing flow fields are not tied to any single source of error and are intended to be broadly applicable across different scenarios where occlusions occur. The interleaved imaging technique also does not address the issue of attenuation with depth. These two sources of error should be addressed separately.

A significant challenge with imaging cardiovascular systems is accurately identifying and tracking the boundary wall motion. The raw ultrasound images were cropped and segmented with thresholding to isolate the boundary walls from the rest of the image, producing a binary segmentation mask that varied over space and time. The flow field data had a different spatial resolution to the image so a nearest-neighbour method was employed to classify the flow field coordinates into a boundary region or fluid region. After the neural field models had denoised the flow fields, the performance had to be evaluated against a ground truth measurement. The dataset collected by Fraser et al. contained scans taken at 2cm depths and 6cm depths under the same flow and ultrasound conditions. This served as a reference for the neural field reconstructions when applied to data that had been subjected to ultrasound attenuation from depth. The dataset also contained many instances of the same flow conditions with different EchoPIV frame rates (different values of $m$), which served as a reference for reconstructing occluded flow fields. Although the reference flow fields were suitable for visual comparison, they could not be considered ground truth. Instead, a similar verification pipeline used by Fraser et al. was employed. The binary segmentation masks were linearly interpolated, assuming the latex tube did not deform significantly during the pulsatile action, and these lines were used as a reference to find the radial positions of the velocity vectors. The flow field was cylindrically integrated to determine the flow rate in m$^3$s$^{-1}$ and then converted to ml min$^{-1}$. This was compared with the flow rate measured by the transit time flow meter used in Fraser et al.'s experiments.

\section{Methods}

After synthetic data generation was complete, four different Fourier Feature encoding strategies were compared on these datasets: no Fourier Features (Vanilla), Random Fourier Features (RFF), Trainable Fourier Features (TFF), and Multi-Scale Fourier Features (MSFF). For this comparison, the model architecture was kept the same with 3 hidden layers of 256 neurons each, and LeakyReLU activation was used. Additionally, the physics-constrained loss function and learning rate was kept the same. This comparison was also performed on the flow field data obtained from real ultrasound scans, collected by Fraser et al. \cite{fraserUltrasoundImagingVelocimetry2017}, to examine the models' transferability and consistency with real-world flow data. After the best encoding strategies were identified, a branched model was created to inpaint occluded regions in the flow field. Although the synthetic datasets had ground truth flow field data, the real EchoPIV flow fields did not, so the reconstructed flow fields were evaluated first against a reference flow field under the same conditions with lower levels of noise, and then against the measured flow rate.

\subsection{Baseline Model and Training}
For the synthetic datasets, four neural field architectures were developed and compared. First, a vanilla neural field with an input layer of 3 nodes, 3 hidden layers of 256 nodes each with LeakyReLU activation, and an output layer of 2 nodes. This baseline model was optimised with a loss function $\mathcal{L}$ that considered data fidelity, boundary conditions, fluid dynamics, and total variation. Given a noisy flow field, the model was trained to reconstruct this flow field but make corrections based on how well the laws of physics are upheld. The data fidelity term was defined as
\begin{equation}
\mathcal{L}_\text{data} = \frac{1}{N}\sum_{i=1}^{N}||\vec{u}(\vec{x}_i) - \vec{u}^*(\vec{x}_i)||_2^2
\end{equation}
where there are in total $N$ collocation points in space and time, $\vec{u}(\cdot)$ is the output of the neural field, and $\vec{u}^*(\cdot)$ is the target output from the dataset. Womersley flow is periodic so cyclic boundary conditions were enforced by comparing the flow field at the first and last timesteps
\begin{equation}
\mathcal{L}_\text{cycle} = \frac{1}{N_\text{cycle}} \sum_{i=1}^{N_\text{cycle}} ||\vec{u}(0, \vec{x}_i) - \vec{u}(T,\vec{x}_i)||_2^2
\end{equation}
where $N_\text{cycle}$ collocation points in space were sampled, and $T$ is the final timestep. Each of the $N$ collocation points in space and time were classified into a fluid region, $\Omega$, or boundary region, $\partial\Omega$. For synthetic data, the boundaries were accurately known, with the first row and last row of the flow fields being the boundary walls. When $\vec{u}$ has no $z$ component, the residual of Equation (\ref{eq:ns1}) evaluated at $\vec{x}$ becomes
\begin{equation}
    R_{NS}(\vec{x}) = \frac{\partial \omega_z}{\partial t} + \frac{\partial}{\partial x}(\omega_z u_x) + \frac{\partial}{\partial y}(\omega_z u_y) - \frac{1}{Re}\nabla^2\omega_z \Bigg{|}_{\vec{x}}
\end{equation}
where $\omega_z = \nabla \times \vec{u}$ is the only non-zero component of the vorticity vector. Thus, the physics loss term was formulated as follows
\begin{equation}
\mathcal{L}_\text{phys} = \frac{1}{|\Omega|} \sum_{\vec{x}_i \in \Omega}R_{NS}(\vec{x}_i)^2 + \frac{1}{|\Omega|} \sum_{\vec{x}_i \in \Omega} (\nabla \cdot \vec{u}(\vec{x}_i))^2 + \frac{1}{|\partial\Omega|}\sum_{\vec{x}_i \in \partial\Omega}||\vec{u}(\vec{x}_i)||_2^2
\end{equation}
where $|\cdot|$ denotes the cardinality of the set, $\vec{\omega}_i = \nabla \times \vec{u}(\vec{x}_i)$ is the vorticity, and $||\cdot||_2^2$ is the L2 norm. When the model in \cite{garziaNeuralFieldsContinuous2024} was applied to real ultrasound scans, its accuracy decreased substantially. The authors hypothesized that this limitation arose from comparing the reconstructed flow only to the velocity field in the first frame, suggesting that a locally comparative approach would be more appropriate for real data. In this work, we address this shortcoming by introducing a total variation loss term, which enforces local spatial consistency in the reconstructed flow fields and thus serves a similar role to the proposed local comparison, potentially improving robustness on real scans. This total variation loss term was defined as
\begin{equation}
\mathcal{L}_\text{TV} = \frac{1}{N} \sum_{i=1}^{N} ||\vec{u}(\vec{x}_i) - \vec{u}(\vec{x}_i+\delta\vec{x})||_2^2
\end{equation}
where $\delta\vec{x}$ is a small finite change in spatial coordinates. The total loss was a weighted sum of these individual loss terms
\begin{equation}
\mathcal{L} = \lambda_\text{data}\mathcal{L}_\text{data} + \lambda_\text{cycle}\mathcal{L}_\text{cycle} + \lambda_\text{phys}\mathcal{L}_\text{phys} + \lambda_\text{TV}\mathcal{L}_\text{TV}
\end{equation}
where $\lambda_\text{data} = \lambda_\text{cycle} = \lambda_\text{TV} = 1$ and $\lambda_\text{phys} \sim 10^{-6}$ are loss weights. Preliminary tests determined the loss weights, supported by an ablation study, based on the order of magnitude with respect to the data fidelity for real ultrasound data. The model parameters were optimised with a learning rate of $10^{-5}$.

\subsection{Fourier Feature Mapping Strategies}
\begin{figure}
    \centering
    \includegraphics[width=0.8\linewidth]{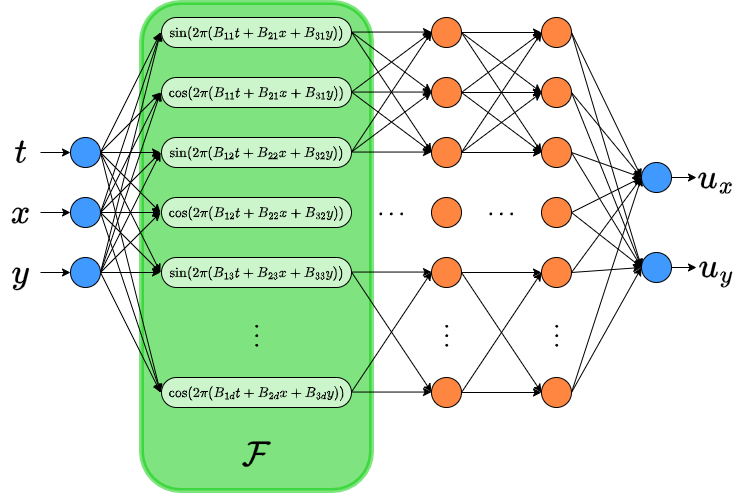}
    \caption{Architecture of the neural fields that used Fourier Features. In the first layer, the spatiotemporal coordinates were projected into a Fourier space so that the subsequent layer of the model had $2d$ input features. For RFF neural fields, $B$ is chosen from a normal distribution with 0 mean and standard deviation 10. For TFF neural fields, $B$ is initiated in this way but was treated as a trainable parameter by the model.}
    \label{fig:nf1}
    \Description[Architecture of a neural field with Fourier Feature encoding.]{Spatiotemporal coordinates are projected into a Fourier space, increasing the number of dimensions for the input of the subsequent layer of the neural network.}
\end{figure}

The baseline Vanilla model was modified to use Random Fourier Features (RFF) proposed by Tancik et al., following the hypothesis that the solutions would be more suitable for flow fields characterised by Fourier series, like the Womersley flow solution in Equation \ref{eq:wom1} \cite{tancikFourierFeaturesLet2020}. This RFF layer, shown in Figure \ref{fig:nf1}, was placed between the input layer and the first hidden layer of the vanilla neural field with Fourier dimension $d=256$ and was applied to both spatial and temporal coordinates, rather than just temporal coordinates as in \cite{garziaNeuralFieldsContinuous2024}. The Fourier modes, $B$, in Equation (\ref{eq:rff1}) were fixed from a normal distribution with standard deviation $\sigma = 10$. 

As an extension to this, we also tested a common adaptation to the RFF model whereby the Fourier modes in $B$ were treated as trainable parameters by the model - this model was called the Trainable Fourier Features (TFF) model. Encouraging the model to choose the most optimal Fourier modes would allow it to adapt to different types of flow and different geometries better than other architectures. It should be noted that Tancik et al. noticed no noticeable improvement by making $B$ trainable when it has the same learning rate as the neural network weights, though their model was applied to extremely high frequency data like natural images and 3D scenes \cite{tancikFourierFeaturesLet2020}. This work tested their claim on flow fields, which sit on a lower dimensional manifold than natural images due to lower frequencies and generalisable physical laws. Further to applying the model to a different type of data, the Fourier modes in $B$ were optimised with a different learning rate ($10^{-2}$) to that used for the network parameters. For high frequency data, it may seem that increasing $\sigma$ would produce a more accurate reconstruction as it provides a larger range of frequencies to the neural field. This high frequency input, however, leads to highly non-smooth loss landscapes making it harder for the model to converge. When combined with trainable Fourier features, the constantly changing inputs for the model only add to this instability which may explain why Tancik et al. saw no obvious advantage. We hypothesise, however, that lower frequency data, like physical fields, may not be subjected to this instability to a severe degree. Moreover, if the desired frequencies lie in a narrow band, random sampling with a large $\sigma$ has a low chance of hitting the right scales - effectively adding feature noise to the model.

Another adaptation we propose is Multi-Scale Fourier Features (MSFF) where the $B$ matrix in RFF is now replaced with a set of matrices $\mathcal{B} = \{B_i\}_{i=1}^{n_B}$, where each $B_i \in \mathbb{R}^{d_i \times n}$ is drawn from a Gaussian distribution with mean 0 and standard deviation $\sigma_i$. Different values of $\sigma$ allow the neural field to capture multi-scale frequencies from low frequency global, smooth trends to high frequency local, fine details. The result of the concatenation of these features produces a non-Gaussian distribution that has peaks at multiple frequency scales and a higher probability of high-frequency components than a single Gaussian. For our experiments $n_B = 3$, $d_i = 85 \quad \forall i$, and $\sigma_i \in \{10,20,40\}$ were chosen.

\subsection{Branched Model}
\begin{figure}[h]
    \centering
    \includegraphics[width=\linewidth]{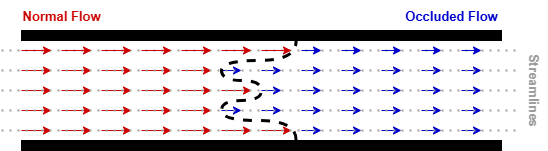}
    \caption{The occlusion map was found by calculating the difference between the velocity at each point along a streamline and the average velocity along that streamline, preserving no-slip boundary conditions. This difference was input into a sigmoid function to produce a probability of a point belonging to the occluded region.}
    \label{fig:occ1}
    \Description[Identifying occluded regions in the flow field.]{The occlusion map was calculated by finding the different between the velocity at each point along streamlines, which were then passed into a sigmoid function to determine the probability that the point belongs to an occluded region.}
\end{figure}

The baseline Vanilla neural field, original RFF neural field, and proposed TFF and MSFF architectures were verified on simulated data to ensure that they were capable of denoising flow fields that may be subject to ultrasound attenuation. These architectures were used to tackle the problem of attenuation with depth. Handling occluded flow fields, however, was a more challenging task. The Fourier Feature-based neural field architectures would likely overfit to the occluded flow field. In contrast, the Vanilla neural field was expected to under-represent the flow field, likely producing a low frequency, constant field. This could act as a prior for inpainting the occluded regions. A new architecture was proposed where the spatiotemporal coordinates were input to both an MSFF neural field and a Vanilla neural field in parallel, with each of them trained to recreate the noisy flow field. A separate neural network would interleave the outputs of these branches based on an occlusion map. To identify occluded regions, it was assumed that the flow should be constant along the length of the tube, so regions where the velocity was lower than the average along a streamline - shown in Figure \ref{fig:occ1} - were recorded and stored in a grid-based occlusion map, $\Lambda \in \mathbb{R}^{\mathcal{T} \times \mathcal{X} \times \mathcal{Y}}$ where $\mathcal{T}$, $\mathcal{X}$, and $\mathcal{Y}$ are the domains of the coordinates. The difference between the velocity in the region and the average velocity along the streamline was calculated and input into a sigmoid function to restrict the output to the range $(0,1)$.

\begin{figure}[h]
    \centering
    \includegraphics[width=\linewidth]{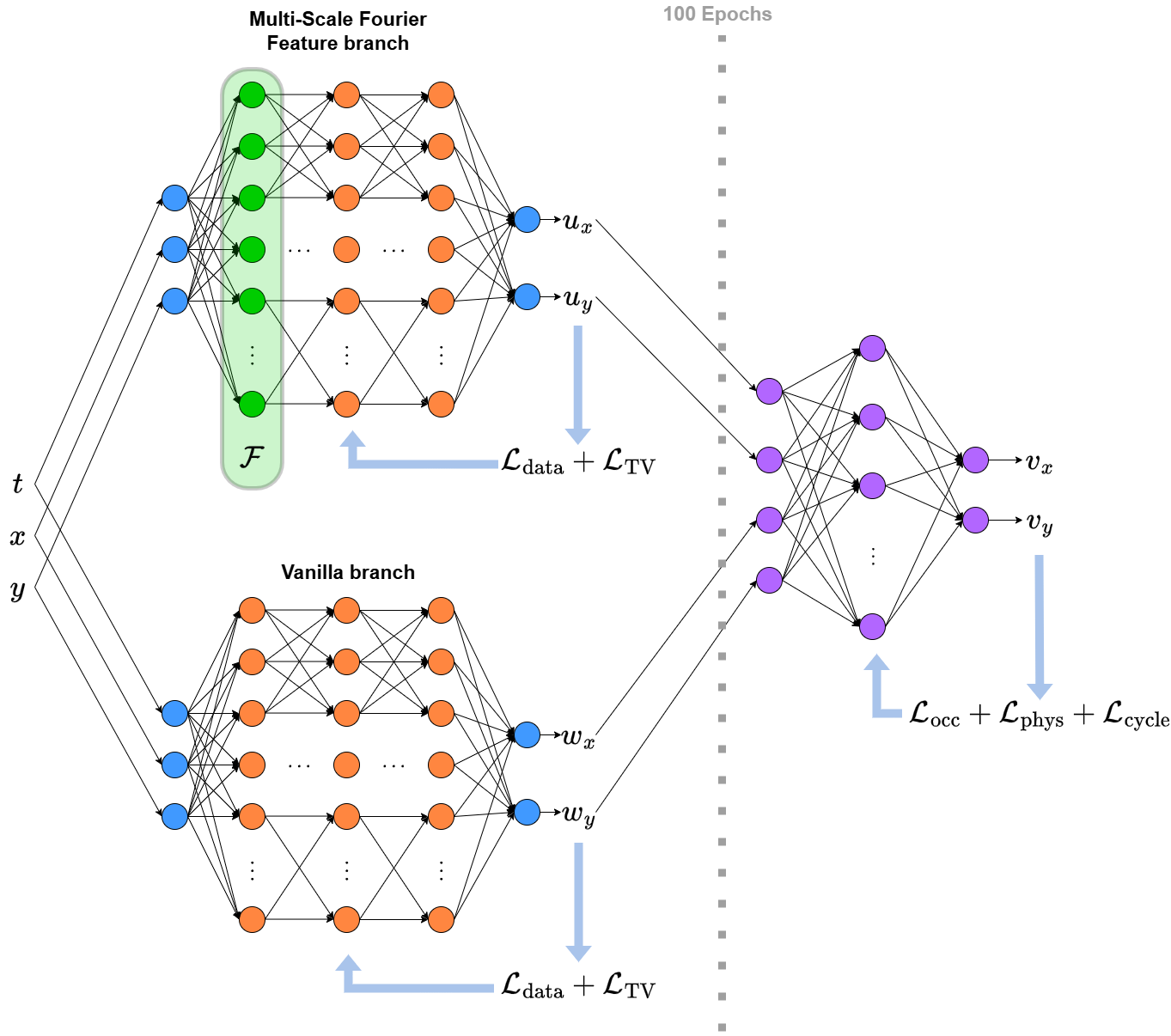}
    \caption{For inpainting occluded regions, the spatiotemporal coordinates were separately fed into two neural field branches: the MSFF model (top) and the Vanilla model (bottom). For the first 100 epochs, these branches were trained to separately optimise data fidelity and total variation. After this, a combining neural network was trained to select the correct branch depending on the coordinate based on the occlusion map $\Lambda$.}
    \label{fig:branch1}
    \Description[Branched model architecture.]{The spatiotemporal coordinates were input separately into a Multi-Scale Fourier Feature neural field and Vanilla neural field. The outputs of these branches were combined with a third neural network that learned the occlusion map.}
\end{figure}

After this first stage of training was complete, the model needed to know which locations to refer to the MSFF model and which locations to refer to the Vanilla model. A smaller neural network was used to combine the outputs of these larger models and was trained with an occlusion loss function
\begin{equation}
    \mathcal{L}_\text{occ} = \frac{1}{N}\sum_{i=1}^{N} ||\vec{v}(\vec{x}_i) - (\Lambda(\vec{x}_i)\vec{u}(\vec{x}_i) + (1-\Lambda(\vec{x}_i))\vec{w}(\vec{x}_i))||^2
\end{equation}
where $N$ is the number of collocation points $\vec{x}_i$ in space and time, $\vec{v}$ is the output of the combining neural network, $\Lambda(\vec{x}_i)$ is the value of the occlusion map at $\vec{x}_i$ that behaves as a probability of not being an occluded region, $\vec{u}$ is the output of the MSFF branch, and $\vec{w}$ is the output of the Vanilla branch. Directly referring to the $\Lambda$ occlusion grid would be faster, but this comes at the cost of restricting the output resolution. The combining neural network preserves the potentially infinite resolution of the prior neural field architectures. A full schematic of the branched architecture is shown in Figure \ref{fig:branch1}.

\section{Results and Discussion}

\subsection{$\lambda_\text{phys}$ Ablation}
An ablation study was performed to determine the value of $\lambda_\text{phys}$ when the other loss weights were set to $1$. The range of values in the study spanned $[10^{-10},1]$. This range was chosen for two reasons: first, the Navier-Stokes equations are 3-dimensional by definition, and out-of-plane motion could not be captured by EchoPIV so the residuals of the Navier-Stokes equations were given a lower weight. Secondly, the classification between the flow region and the boundary region for the real ultrasound data had uncertainty that would carry over to $\mathcal{L}_\text{phys}$. A lower weight would reduce the effect of the propagation of this uncertainty. 

\begin{figure}[h!]
    \centering
    \includegraphics[width=\linewidth]{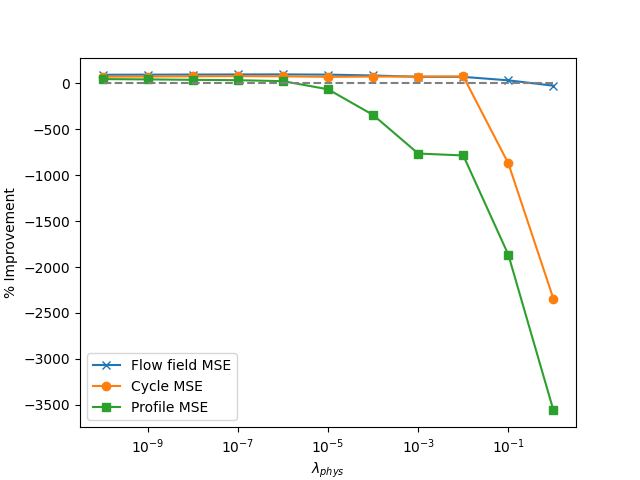}
    \caption{The percentage reduction in flow field MSE, cycle MSE, and profile MSE when the RFF architecture was applied to a synthetic dataset with a moderate level noise (exp12) for different values of $\lambda_\text{phys}$. The model provides no improvement after $\lambda_\text{phys} \sim 10^{-6}$}.
    \label{fig:ablation}
    \Description[Results of ablation for $\lambda_\text{phys}$]{When considering flow field MSE, cycle MSE, and profile MSE, the RFF architecture provided less improvement when applied to synthetic data as the physics loss parameter was increased.}
\end{figure}

A synthetic dataset with a moderate level of noise was chosen, and the RFF architecture was trained to reconstruct this flow field after 50 epochs of training. In addition to comparing the reconstructed flow field with the original Womersley flow simulation, we also compared the average flow velocity over time to ensure that the simulated cardiac cycle was also reconstructed. The physics loss term, $L_\text{phys}$ also includes terms that enforce no-slip boundary conditions, which gives rise to a radially dependent profile that is characteristic of Womersley flow \cite{womersleyMethodCalculationVelocity1955}. Due to this, the ablation study also compared the reconstructed flow profile with the analytic solution. Figure \ref{fig:ablation} shows that when $\lambda_\text{phys} > 10^{-6}$, the model performs worse than the original noisy dataset. There is a general trend for the percentage reduction in MSE to decrease as $\lambda_\text{phys}$ increases. However, a lower value of $\lambda_\text{phys}$, would give less importance to enforcing the physical laws. Thus, the highest value of $\lambda_\text{phys}$ that caused a decrease in all comparison metrics was when $\lambda_\text{phys} = 10^{-6}$.

\subsection{Synthetic Data}
\begin{figure}[h]
    \begin{subfigure}[b]{.45\linewidth}
        \includegraphics[width=\linewidth]{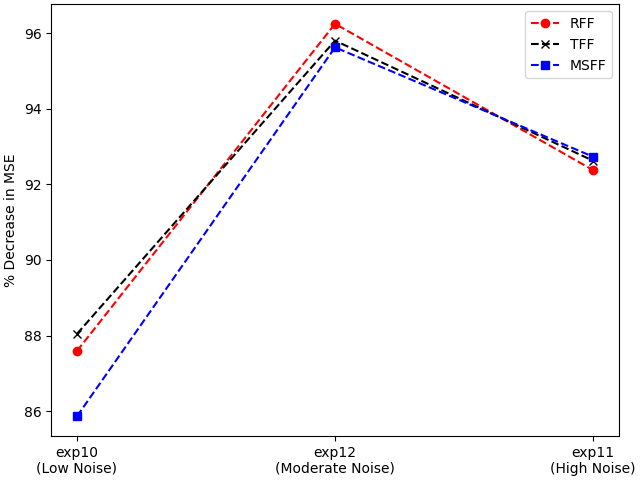}
        \caption{}\label{fig:low_alpha}
    \end{subfigure}
    \begin{subfigure}[b]{.45\linewidth}
        \includegraphics[width=\linewidth]{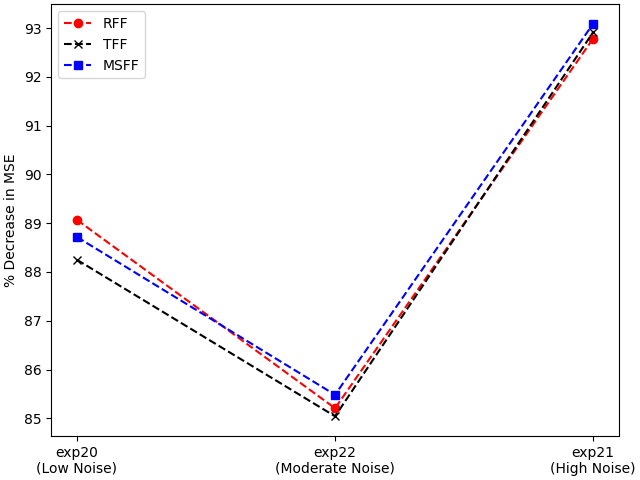}
        \caption{}\label{fig:high_alpha}
    \end{subfigure}
    \caption{Relative decrease in MSE for the three Fourier feature based models when applied to datasets with (a) $\alpha=2.77$ and (b) $\alpha=5$. No one model was consistently better than the others and they all had comparable performance, though the MSFF model produced the largest improvement for very noisy and high frequency data.}
    \label{fig:alpha_results}
    \Description[Results of Fourier Feature models on synthetic data]{As the level of noise increased, the performance of the models decreased independent of Womersley number.}
\end{figure}

\begin{figure}[h]
\begin{subfigure}[b]{.45\linewidth}
\includegraphics[width=\linewidth]{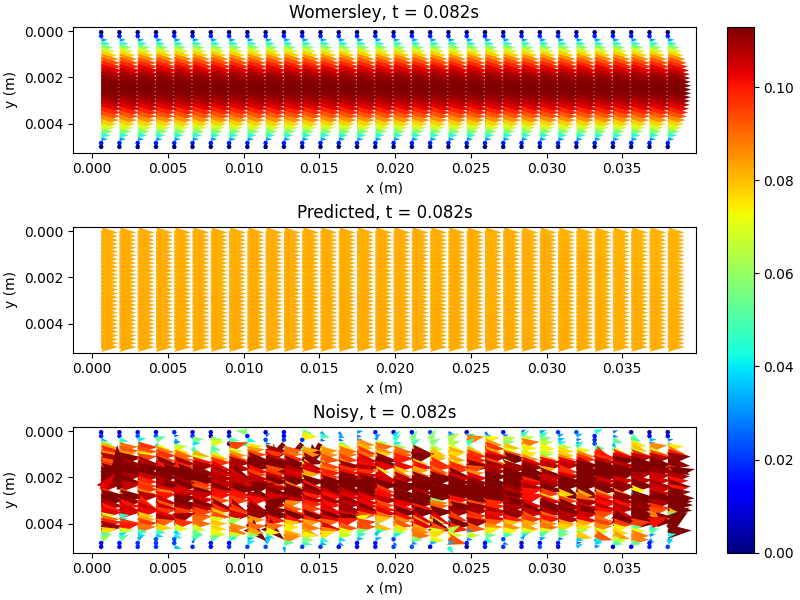}
\caption{}\label{fig:exp12_van_flow1}
\end{subfigure}
\begin{subfigure}[b]{.45\linewidth}
\includegraphics[width=\linewidth]{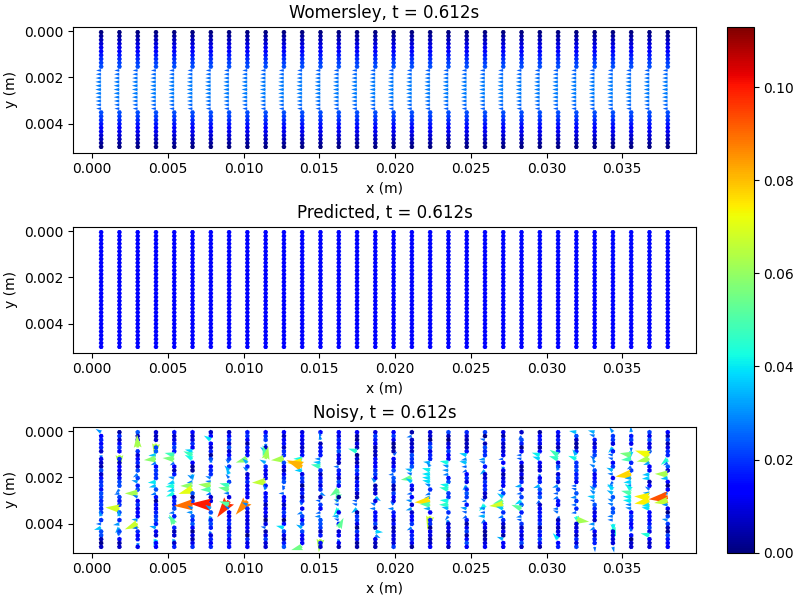}
\caption{}\label{fig:exp12_van_flow2}
\end{subfigure}
\begin{subfigure}[b]{.45\linewidth}
\includegraphics[width=\linewidth]{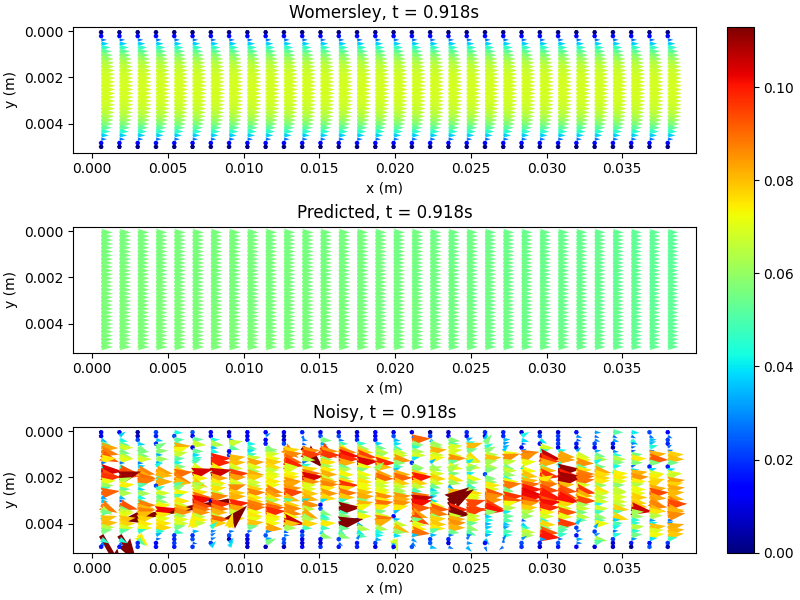}
\caption{}\label{fig:exp12_van_flow3}
\end{subfigure}
\begin{subfigure}[b]{.45\linewidth}
\includegraphics[width=\linewidth]{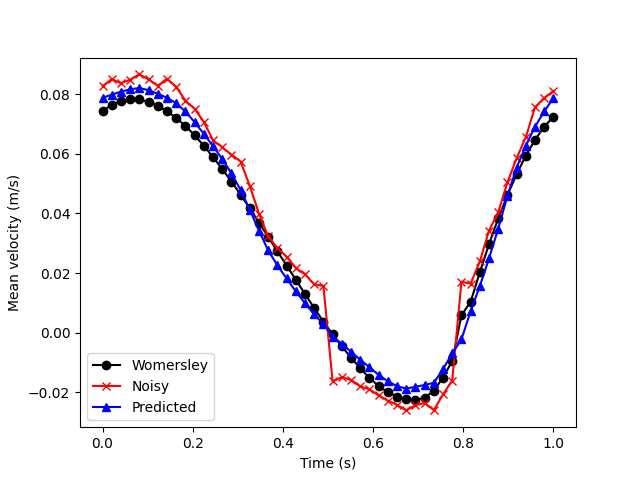}
\caption{}\label{fig:exp12_van_cycle}
\end{subfigure}
\caption{Qualitative results from applying the vanilla neural field to the exp12 dataset. The flow fields of the analytic solution (top), prediction from the neural field model (middle), and the noisy synthetic dataset (bottom) are shown at times (a) $t=0.082$s, (b) $t=0.612$s, and (c) $t=0.918$s. (d) compares how the mean velocity changes over time between the three datasets. The mean velocity cycle is closer to the analytic solution than the original noisy dataset, but the flow fields do not vary spatially - only temporally.}
\label{fig:exp12_van}
\Description[Results of Vanilla Neural Field applied to moderately noisy synthetic dataset with low Womersley number.]{Predicted flow field at various points in time do not vary spatially, despite the mean velocity over time being in good agreement with the analytic Womersley flow.}
\end{figure}

\begin{figure}[h]
\begin{subfigure}[b]{.45\linewidth}
\includegraphics[width=\linewidth]{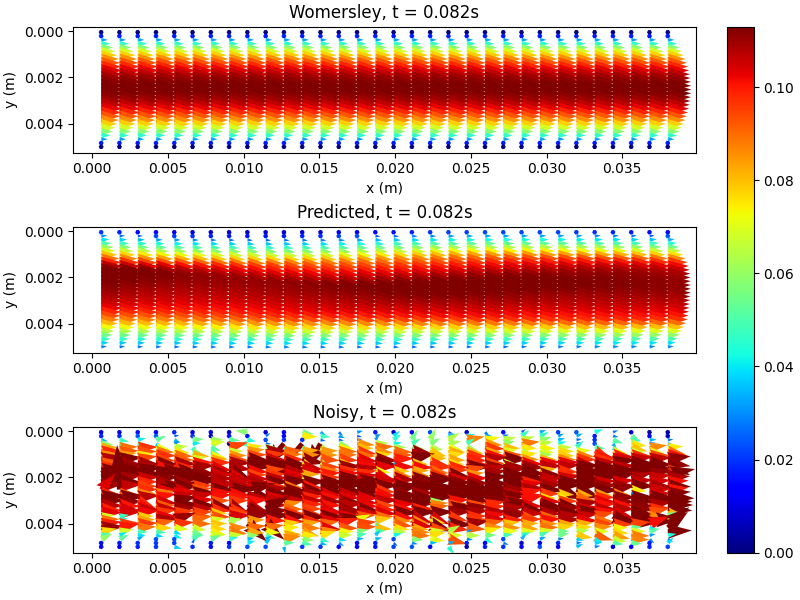}
\caption{}\label{fig:exp12_rff_flow1}
\end{subfigure}
\begin{subfigure}[b]{.45\linewidth}
\includegraphics[width=\linewidth]{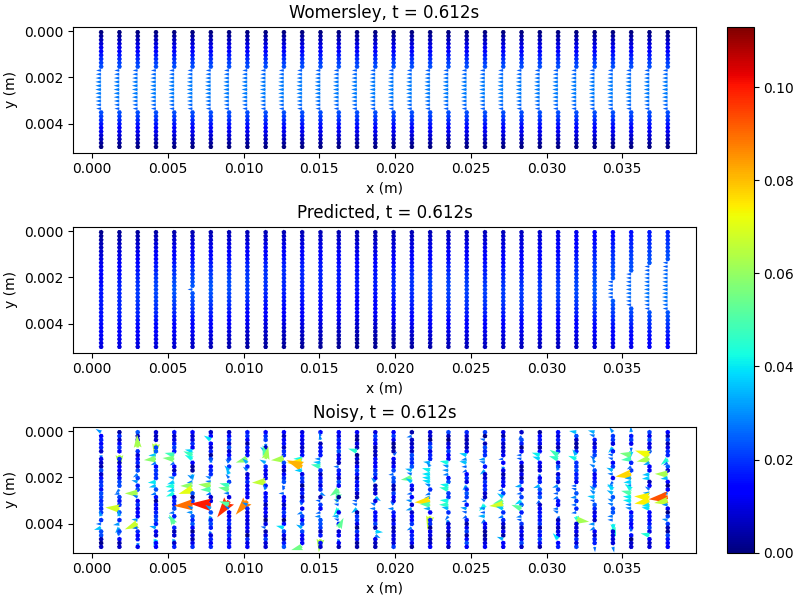}
\caption{}\label{fig:exp12_rff_flow2}
\end{subfigure}
\begin{subfigure}[b]{.45\linewidth}
\includegraphics[width=\linewidth]{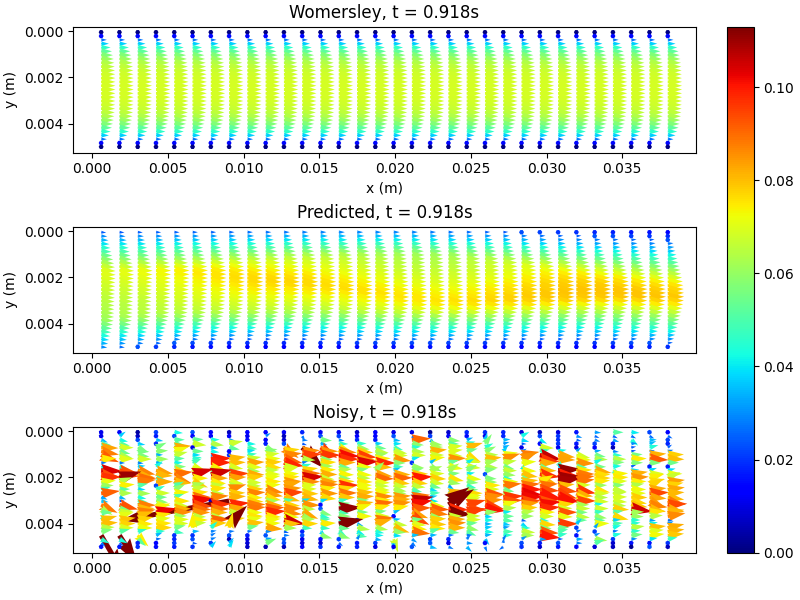}
\caption{}\label{fig:exp12_rff_flow3}
\end{subfigure}
\begin{subfigure}[b]{.45\linewidth}
\includegraphics[width=\linewidth]{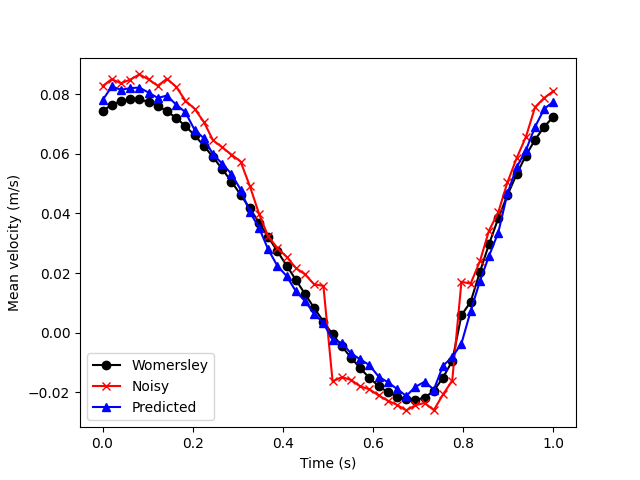}
\caption{}\label{fig:exp12_rff_cycle}
\end{subfigure}
\caption{Qualitative results from applying the RFF neural field to the exp12 dataset. The flow fields of the analytic solution (top), prediction from the neural field model (middle), and the noisy synthetic dataset (bottom) are shown at times (a) $t=0.082$s, (b) $t=0.612$s, and (c) $t=0.918$s. (d) compares how the mean velocity changes over time between the three datasets. The flow field is now varying spatially as well as temporally due to the inclusion of Fourier features, while keeping the mean velocity dynamics close to those of the analytic solution. }
\label{fig:exp12_rff}
\Description[Results of Random Fourier Features applied to moderately noisy synthetic dataset with low Womersley number.]{Predicted flow field now varies spatially as well as temporally, with no slip boundary conditions met.}
\end{figure}

\begin{figure}[h]
\begin{subfigure}[b]{.45\linewidth}
\includegraphics[width=\linewidth]{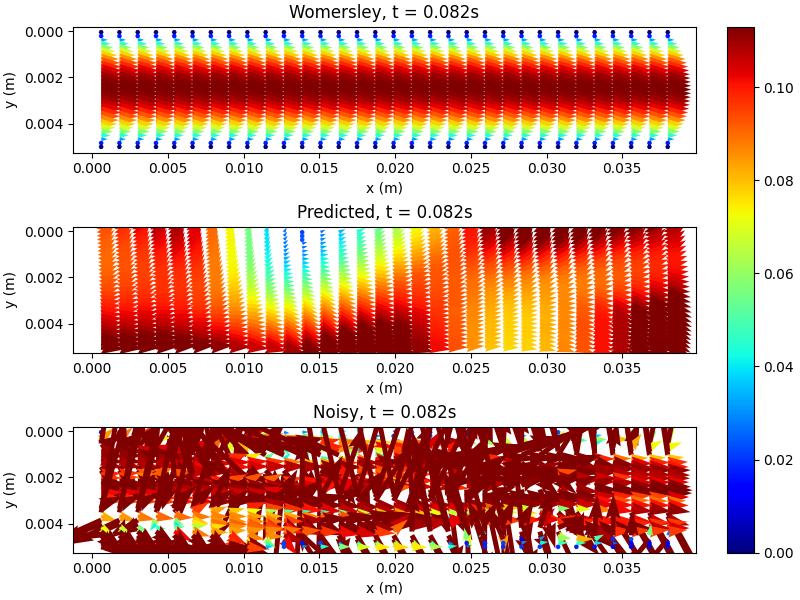}
\caption{}\label{fig:exp11_msff_flow1}
\end{subfigure}
\begin{subfigure}[b]{.45\linewidth}
\includegraphics[width=\linewidth]{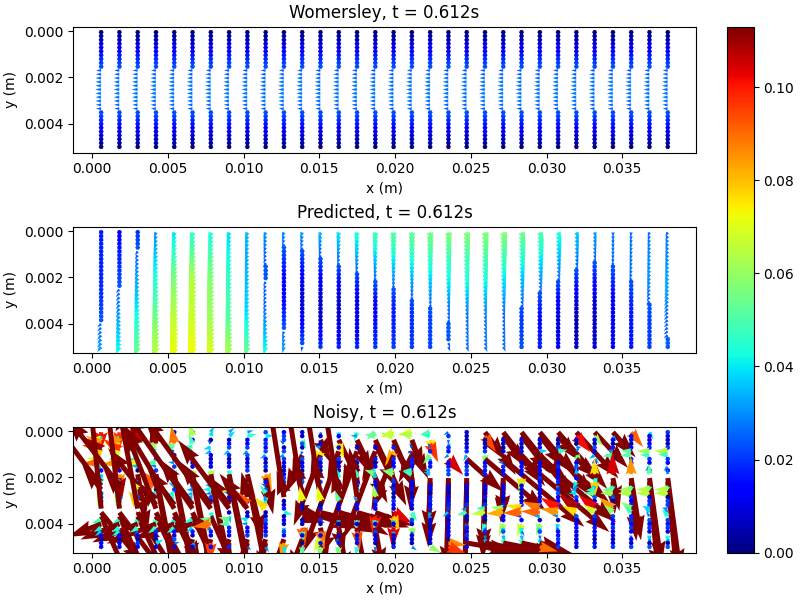}
\caption{}\label{fig:exp11_msff_flow2}
\end{subfigure}
\begin{subfigure}[b]{.45\linewidth}
\includegraphics[width=\linewidth]{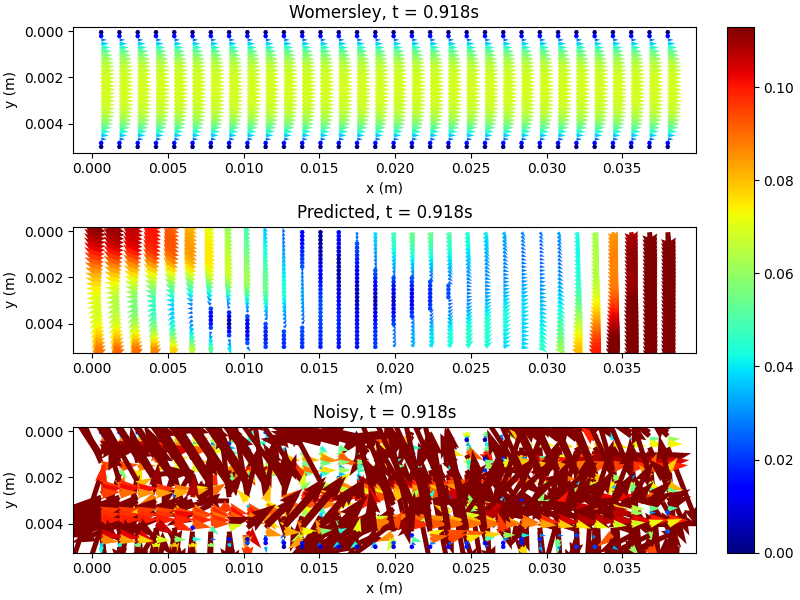}
\caption{}\label{fig:exp11_msff_flow3}
\end{subfigure}
\begin{subfigure}[b]{.45\linewidth}
\includegraphics[width=\linewidth]{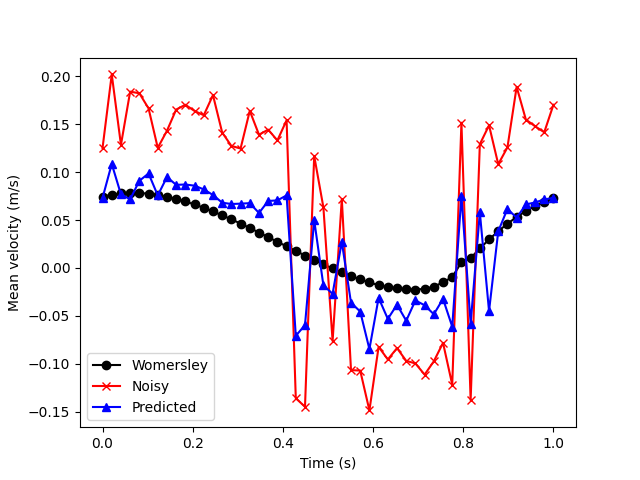}
\caption{}\label{fig:exp11_msff_cycle}
\end{subfigure}
\caption{Qualitative results from applying the MSFF neural field to the exp11 dataset. The flow fields of the analytic solution (top), prediction from the neural field model (middle), and the noisy synthetic dataset (bottom) are shown at times (a) $t=0.082$s, (b) $t=0.612$s, and (c) $t=0.918$s. (d) compares how the mean velocity changes over time between the three datasets. The predicted flow field is smoother, though unphysical. Despite this, the mean velocity cycle is closer to the analytic solution.}
\label{fig:exp11_msff}
\Description[Results of Multi-Scale Fourier Features applied to extremely noisy synthetic dataset with low Womersley number.]{Predicted flow fields do not resemble ground truth despite increased smoothness, though the mean velocity over time is closer to analytic solution.}
\end{figure}

All four neural field models were tested on synthetic datasets with different levels of noise that simulated the conditions of arterial flow with Womersley number $\alpha = 2.77$. Although all four models saw a reduction in the MSE compared to the analytic Womersley flow, Figure \ref{fig:exp12_van} shows an example of how the vanilla neural field models produced constant fields in space that oscillated in time. This lack of spatial variation, which was likely due to the aforementioned spectral bias of neural networks, deemed the vanilla neural field unsuitable for application to real data despite the low MSE. For $\alpha = 2.77$, Figure \ref{fig:low_alpha} shows that the TFF model produced a slightly lower MSE than the RFF model most of the time and Figure \ref{fig:exp12_rff} shows that the inclusion of Fourier features allows the predicted flow field to vary spatially and temporally, while also predicting a mean velocity cycle that is closer to the analytic solution. All models struggled with the extreme levels of noise in the exp11 dataset, though the MSFF model produced the lowest MSE for this dataset, shown in Figure \ref{fig:exp11_msff}. Although the predicted flow fields are unphysical for the system, the mean velocity cycle is closer to the analytic solution so the model has succeeded in denoising the flow, just on a broader scale. 

\begin{figure}[h]
\begin{subfigure}[b]{.45\linewidth}
\includegraphics[width=\linewidth]{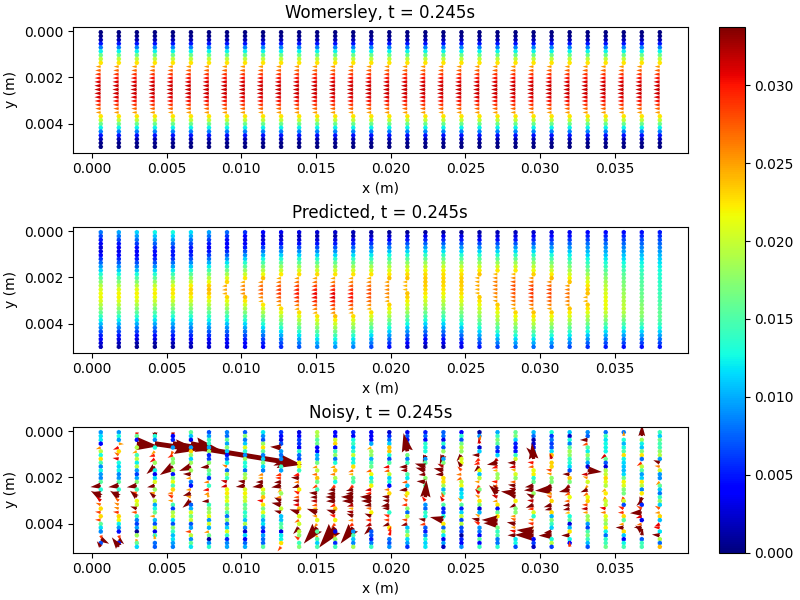}
\caption{}\label{fig:exp20_rff_flow1}
\end{subfigure}
\begin{subfigure}[b]{.45\linewidth}
\includegraphics[width=\linewidth]{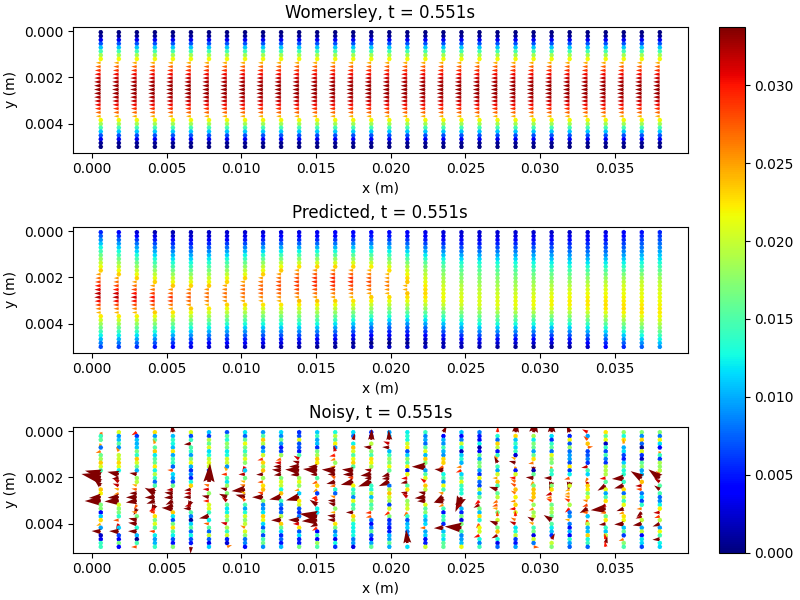}
\caption{}\label{fig:exp20_rff_flow2}
\end{subfigure}
\begin{subfigure}[b]{.45\linewidth}
\includegraphics[width=\linewidth]{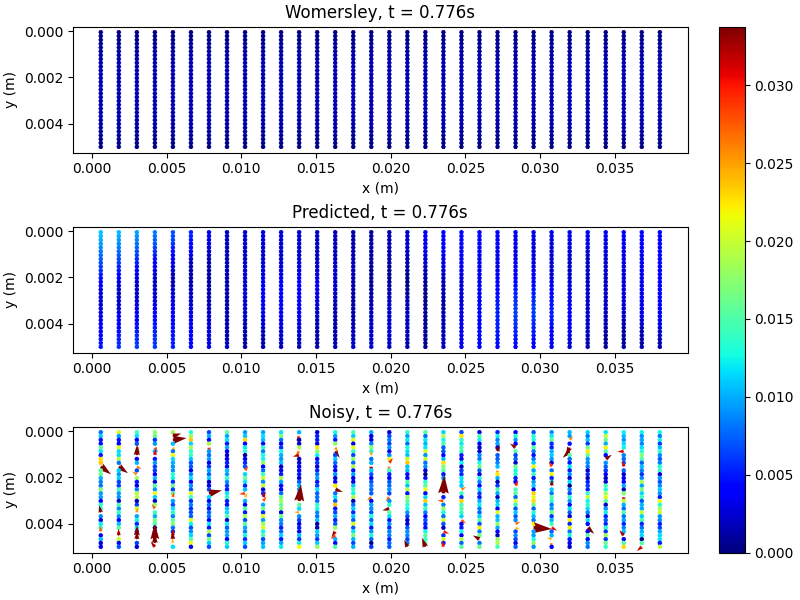}
\caption{}\label{fig:exp20_rff_flow3}
\end{subfigure}
\begin{subfigure}[b]{.45\linewidth}
\includegraphics[width=\linewidth]{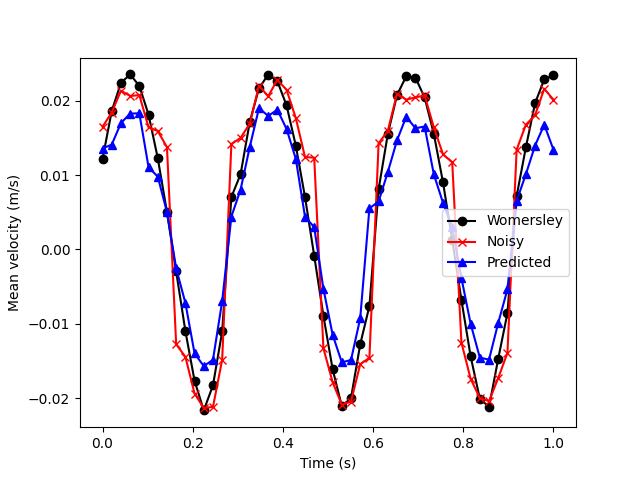}
\caption{}\label{fig:exp20_rff_cycle}
\end{subfigure}
\caption{Qualitative results from applying the RFF neural field to the exp20 dataset. The flow fields of the analytic solution (top), prediction from the neural field model (middle), and the noisy synthetic dataset (bottom) are shown at times (a) $t=0.245$s, (b) $t=0.551$s, and (c) $t=0.776$s. (d) compares how the mean velocity changes over time between the three datasets. Parabolic profile is retained throughout the length of the tube and the predicted cycle agrees with the analytic solution, though the velocities appear to be underestimated.}
\label{fig:exp20_rff}
\Description[Results of Random Fourier Features applied to low noise synthetic dataset with high Womersley number.]{The predicted flow fields show close resemblance to the ground truth, though the flow is not as smooth as expected. Predicted mean velocity over time is slightly lower than the ground truth, though the frequency of the cycle is correct.}
\end{figure}

\begin{figure}[h]
\begin{subfigure}[b]{.45\linewidth}
\includegraphics[width=\linewidth]{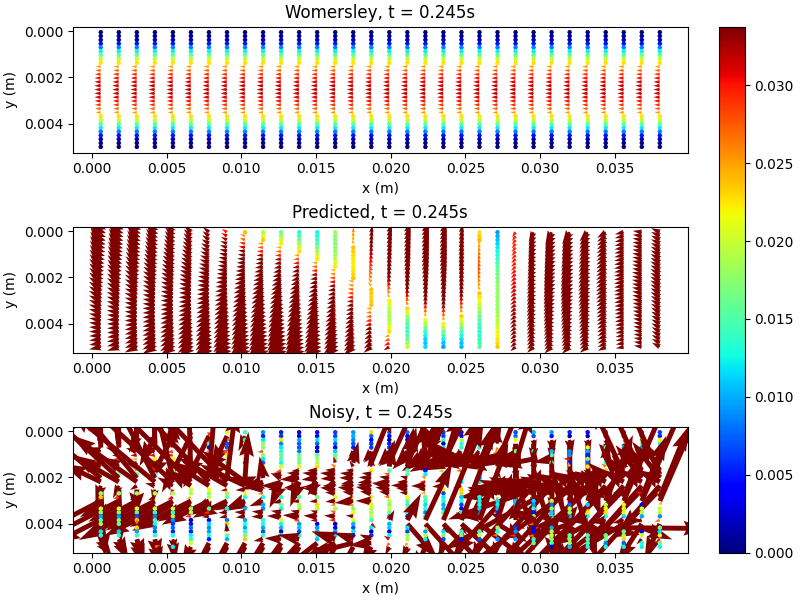}
\caption{}\label{fig:exp21_msff_flow1}
\end{subfigure}
\begin{subfigure}[b]{.45\linewidth}
\includegraphics[width=\linewidth]{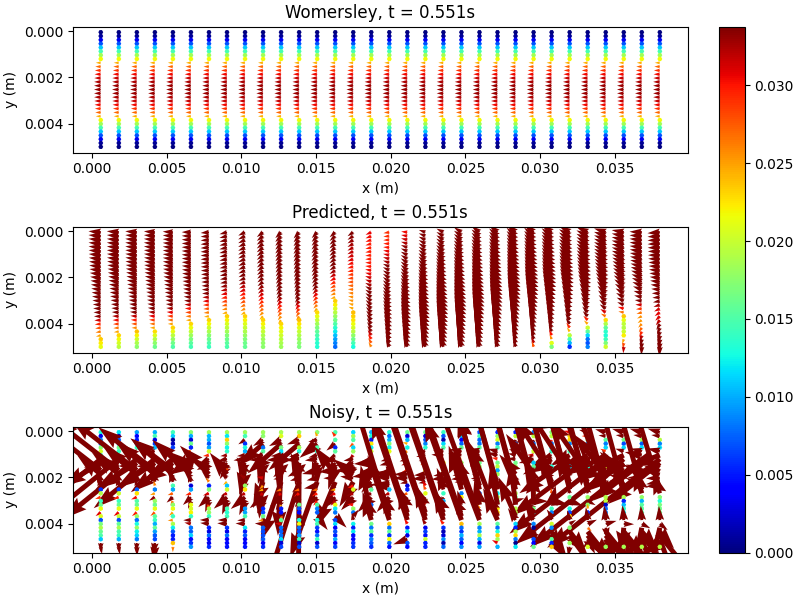}
\caption{}\label{fig:exp21_msff_flow2}
\end{subfigure}
\begin{subfigure}[b]{.45\linewidth}
\includegraphics[width=\linewidth]{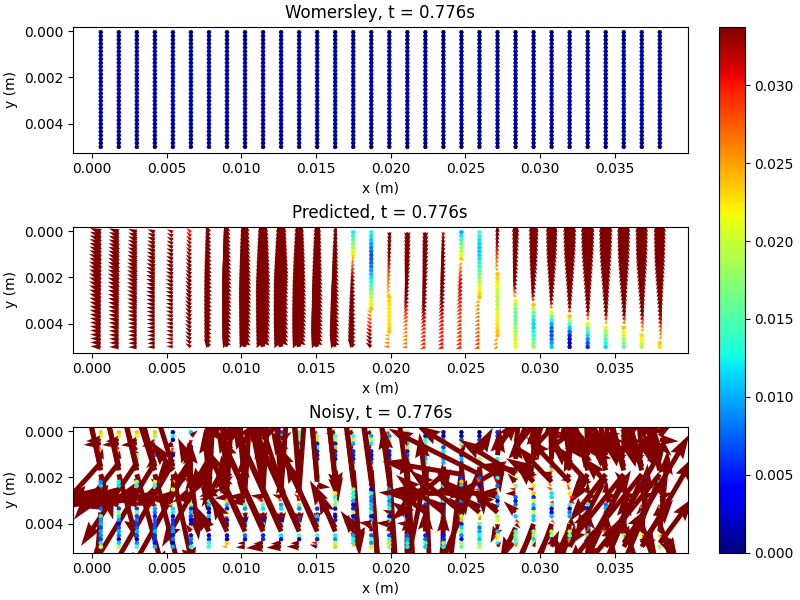}
\caption{}\label{fig:exp21_msff_flow3}
\end{subfigure}
\begin{subfigure}[b]{.45\linewidth}
\includegraphics[width=\linewidth]{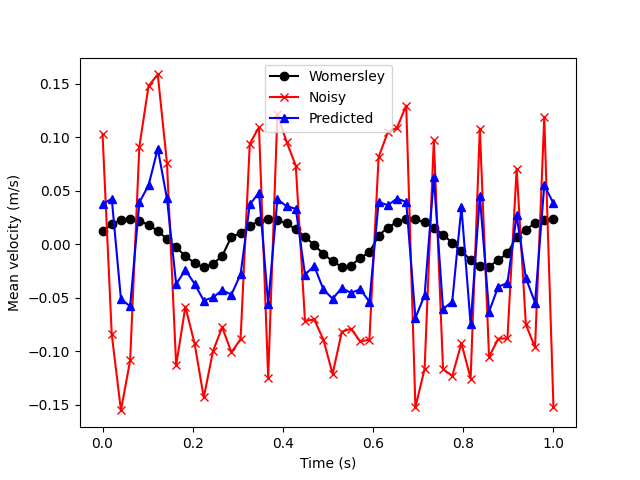}
\caption{}\label{fig:exp21_msff_cycle}
\end{subfigure}
\caption{Qualitative results from applying the MSFF neural field to the exp21 dataset. The flow fields of the analytic solution (top), prediction from the neural field model (middle), and the noisy synthetic dataset (bottom) are shown at times (a) $t=0.245$s, (b) $t=0.551$s, and (c) $t=0.776$s. (d) compares how the mean velocity changes over time between the three datasets. No significant flow features were retained, however the predicted flow cycle, though not in phase with the analytic solution, is closer to it than the original flow field.}
\label{fig:exp21_msff}
\Description[Results of Multi-Scale Fourier Features on extremely noisy synthetic dataset with high Womersley number.]{Predicted flow fields are smooth but are very unphysical with high velocities near boundary walls. Predicted mean velocity over time is higher than expected and frequency is not constant.}
\end{figure}

\begin{figure}[h]
\begin{subfigure}[b]{.45\linewidth}
\includegraphics[width=\linewidth]{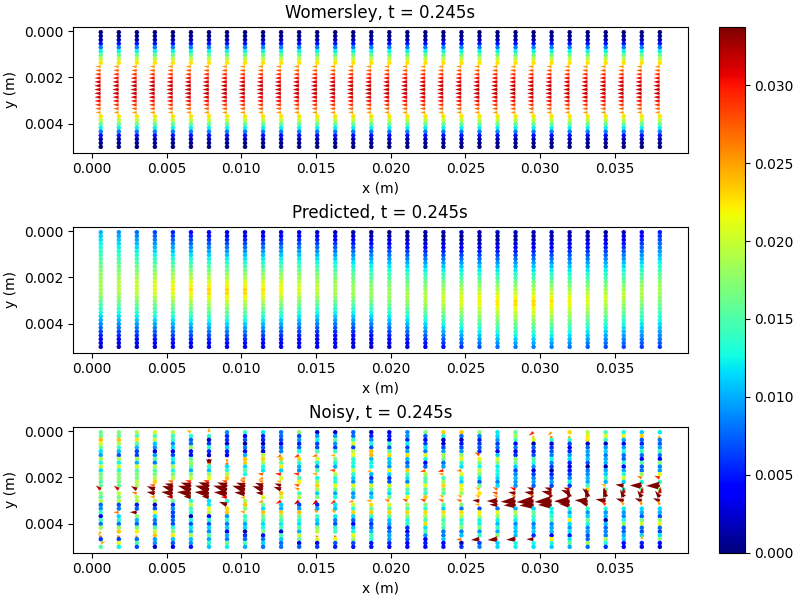}
\caption{}\label{fig:exp22_msff_flow1}
\end{subfigure}
\begin{subfigure}[b]{.45\linewidth}
\includegraphics[width=\linewidth]{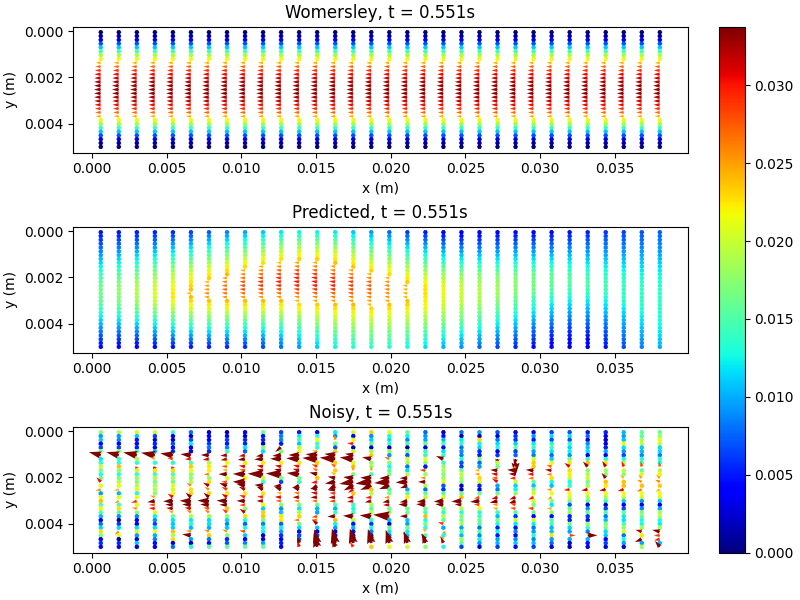}
\caption{}\label{fig:exp22_msff_flow2}
\end{subfigure}
\begin{subfigure}[b]{.45\linewidth}
\includegraphics[width=\linewidth]{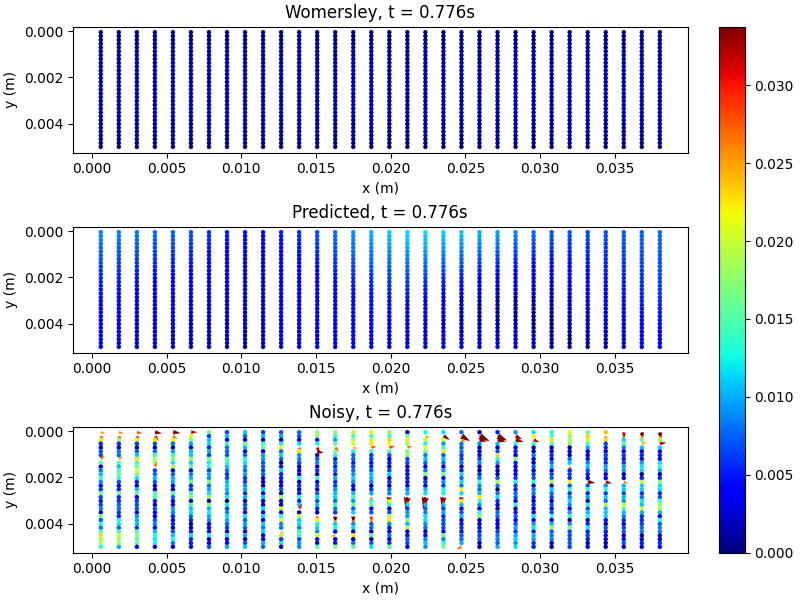}
\caption{}\label{fig:exp22_msff_flow3}
\end{subfigure}
\begin{subfigure}[b]{.45\linewidth}
\includegraphics[width=\linewidth]{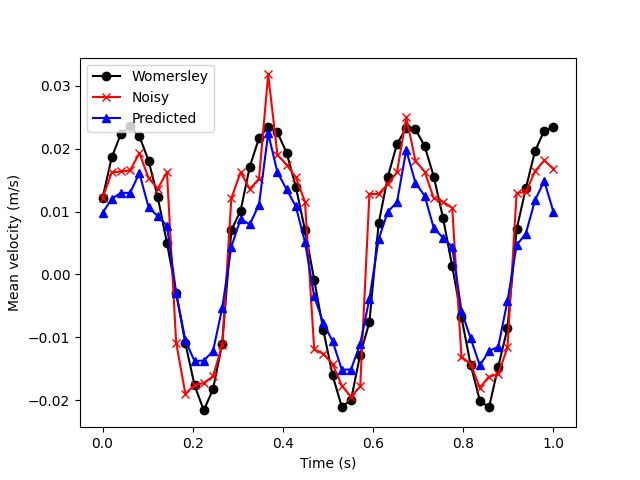}
\caption{}\label{fig:exp22_msff_cycle}
\end{subfigure}
\caption{Qualitative results from applying the MSFF neural field to the exp22 dataset. The flow fields of the analytic solution (top), prediction from the neural field model (middle), and the noisy synthetic dataset (bottom) are shown at times (a) $t=0.245$s, (b) $t=0.551$s, and (c) $t=0.776$s. (d) compares how the mean velocity changes over time between the three datasets. Parabolic profile is retained, but does not have a constant width along the length of the tube. Predicted mean velocity cycle has moved outliers closer to the analytic solution.}
\label{fig:exp22_msff}
\Description[Results of Multi-Scale Fourier Features applied to moderately noisy synthetic dataset with high Womersley number.]{Predicted flow field is not as smooth as ground truth but follows no slip boundary conditions. Mean velocity over time is underestimated but frequency of pulsing flow is in agreement with ground truth.}
\end{figure}

To further test which adaptation of the neural field would be best for particularly chaotic data that is likely to occur with ultrasound data, the experiments were repeated with three more datasets with $\alpha = 5$. Increasing the Womersley number increased the frequency of the flow in both the spatial and temporal domains. This served as a simulation of ultrasound dropout, where the frame rate of the ultrasound system is not high enough to capture rapidly changing flows well. When applied to the higher frequency datasets, Figure \ref{fig:high_alpha} shows that the TFF model saw no significant advantage, while the MSFF model produced the lowest MSE while still retaining flow features. Figure \ref{fig:exp21_msff} shows that the level of noise in the exp21 dataset may be too high for the MSFF model to retain significant flow features. Similar to when the models were applied to the exp11 dataset, the MSFF model denoised the mean velocity cycle but could not retain flow features. In contrast, Figure \ref{fig:exp22_msff} shows that the MSFF model managed to retain the parabolic profile of the flow when applied to datasets with less noise, like exp22. For the exp20 dataset (that had the lowest level of noise out of the datasets with a higher Womersley number), the RFF model predicted a flow field with the lowest MSE and, as shown in Figure \ref{fig:exp20_rff}, more of the parabolic profile is retained. As with the datasets where $\alpha = 2.77$, the vanilla neural field produced a constant field in space that only oscillates in time, similar to Figure \ref{fig:exp12_van}. The models do not perform as well on the datasets where $\alpha = 5$ as they do on those with $\alpha = 2.77$ due to the higher frequency data, which can often be confused with noise, making it harder for the models to discern underlying flow patterns.

In summary, all three Fourier Feature models outperformed the vanilla neural field architecture, as expected. Due to the vanilla neural field model not capturing any spatial resolution, it was not applied to the real ultrasound datasets. Given that the real ultrasound data contains more complex boundary dynamics, it was expected that the MSFF model would perform quite well due to its flexible frequency range. Figure \ref{fig:alpha_results} shows a general trend where the RFF model outperforms others for datasets with lower levels of noise, the TFF model outperforms others for datasets with moderate levels of noise, and the MSFF models prospers with very noisy data. This implies that a single model may not be the way forward and, instead, the most suitable model depends on the level of noise in the dataset. With no exact ground truth flow field to compare with the real ultrasound data, it is difficult to quantify the level of noise to determine the best model architecture to use. The main challenge would be adapting these models that have performed well in idealistic conditions to the unpredictable shortfalls of real data.

\subsection{Denoising Real Ultrasound Attenuation}

The issue of noise due to ultrasound attenuation with depth was addressed first. Phantom E was scanned at both a 2cm and 6cm depth, which were considered reference and noisy datasets, respectively. For each scan, two different interleaved modes were used: $m=31$ and $m=128$ (conventional ultrasound). The RFF, TFF, MSFF, and newer Branched neural field architectures were applied to both datasets and compared with the reference.

\begin{figure}[h]
    \begin{subfigure}[b]{.45\linewidth}
        \includegraphics[width=\linewidth]{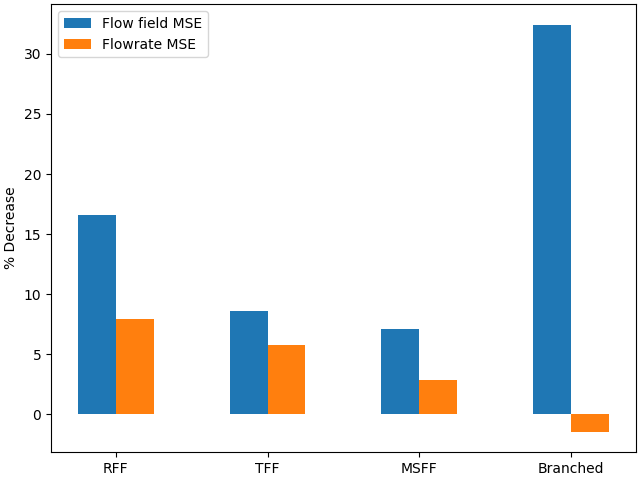}
        \caption{}\label{fig:m31}
    \end{subfigure}
    \begin{subfigure}[b]{.45\linewidth}
        \includegraphics[width=\linewidth]{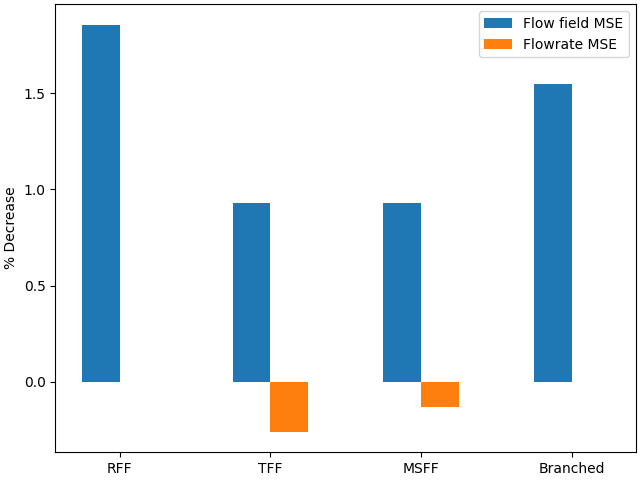}
        \caption{}\label{fig:m128}
    \end{subfigure}
    \caption{Relative decrease in flow field and flowrate MSE for the 4 Fourier feature based models when applied to data subjected to attenuation with depth for (a) $m=31$ and (b) $m=128$. For each value of $m$, the scan collected at 6cm depth was denoised and compared with that collected at 2cm depth.}
    \label{fig:depth_results}
    \Description[Results of 4 Fourier Feature models applied to real data subjected to attenuation with depth]{First bar chart shows that, for a higher frame rate, the branched model reconstructs the flow field best but has worst agreement of flow rate with the experimental measurements. Second bar chart shows that, for lower frame rates, Random Fourier Features and the branched model had the best flow field reconstructions, but no model improved the flow rate mean-squared error.}
\end{figure}

\begin{figure}[h]
\begin{subfigure}[b]{.45\linewidth}
\includegraphics[width=\linewidth]{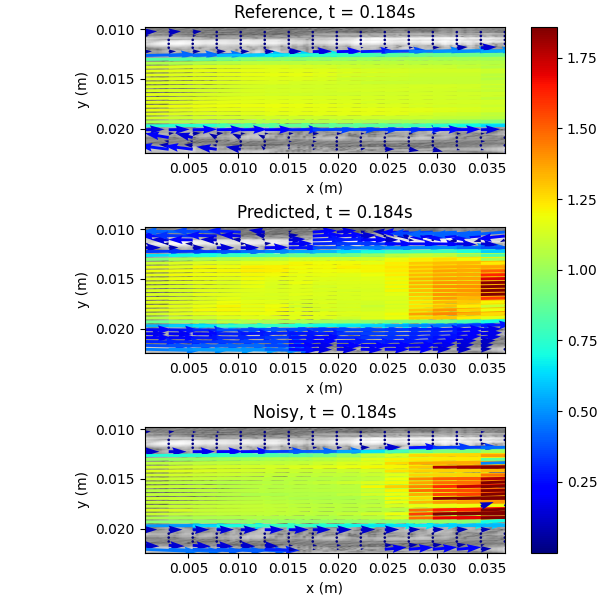}
\caption{}\label{fig:6cm_e31_flow1}
\end{subfigure}
\begin{subfigure}[b]{.45\linewidth}
\includegraphics[width=\linewidth]{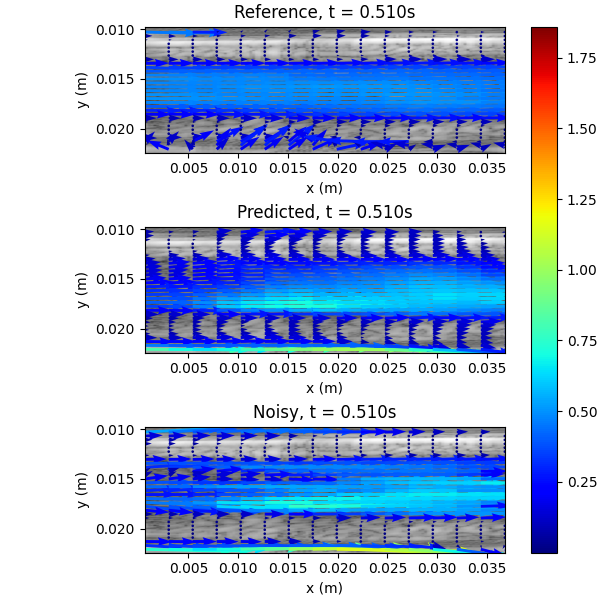}
\caption{}\label{fig:6cm_e31_flow2}
\end{subfigure}
\begin{subfigure}[b]{.45\linewidth}
\includegraphics[width=\linewidth]{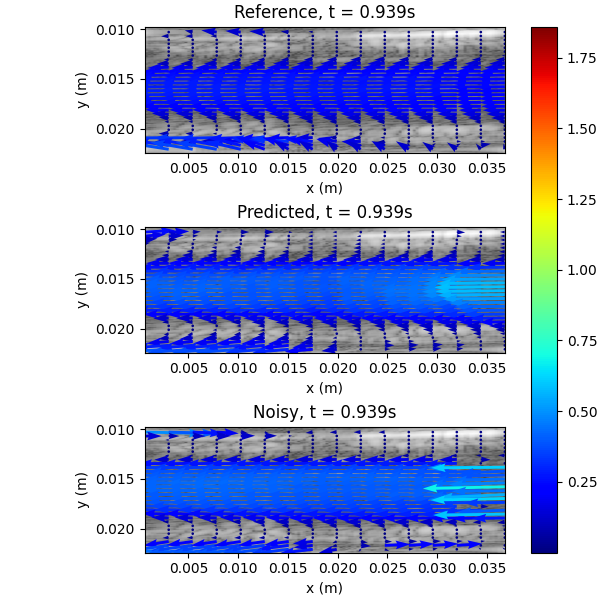}
\caption{}\label{fig:6cm_e31_flow3}
\end{subfigure}
\begin{subfigure}[b]{.45\linewidth}
\includegraphics[width=\linewidth]{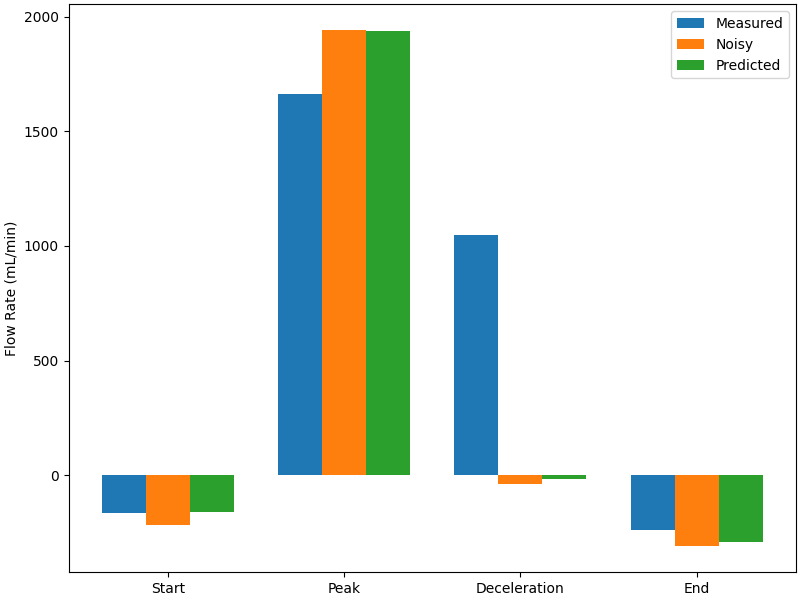}
\caption{}\label{fig:6cm_e31_flowrate}
\end{subfigure}
\caption{Qualitative results from applying the RFF neural field to the scan of phantom E with $m=31$ at a depth of 6cm. The flow fields of the reference at 2cm depth (top), prediction from the neural field model (middle), and the noisy dataset (bottom) are shown at times (a) $t=0.184$s, (b) $t=0.510$s, and (c) $t=0.918$s. (d) compares the calculated flow rate with the measured flow rate. The predicted flow field smoothed out anomalies at the edges of the frame.}
\label{fig:6cm_e31}
\Description[Results of Random Fourier Features applied to scan of phantom E at depth of 6 centimetres with high framerate.]{Predicted flow field is in good agreement with reference but some features from the noisy flow field that should not exist have been retained in the reconstruction. These features are mostly small regions with abnormally high velocities that do not appear in the reference data. Predicted flow rate is in good agreement with experimental measurements at all stages of cardiac cycle except for the deceleration stage.}
\end{figure}

Figure \ref{fig:m31} shows that, when compared with the corresponding scans taken at a depth of 2cm, the Branched model produced the lowest MSE when $m=31$. In contrast, the RFF model produced the lowest MSE when comparing the calculated flow rate to the measured flow rate. In Figure \ref{fig:6cm_e31}, the predicted flow field is smoother than the original noisy flow field. More specifically, the most common regions where noise was observed were at the edges of the ultrasound transducer (the right and left edges of the flow fields). In contrast, the simulated noise in the synthetic data was consistent across the length of the tube which may explain why the TFF and MSFF models did not perform as expected on the real data; these models may have been too complex. The branched model, though containing the MSFF architecture, also uses the Vanilla architecture allowing the model to capture a larger range of frequencies than the MSFF model alone, promoting versatility across all levels of noise.

\begin{figure}[h]
\begin{subfigure}[b]{.45\linewidth}
\includegraphics[width=\linewidth]{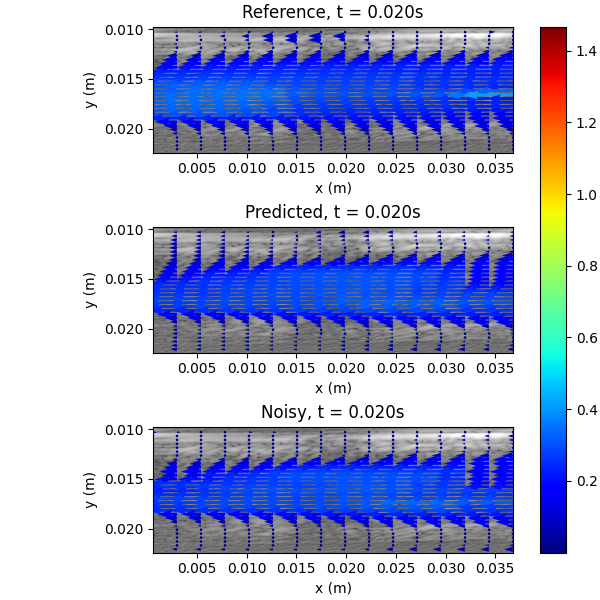}
\caption{}\label{fig:6cm_e_conv_flow1}
\end{subfigure}
\begin{subfigure}[b]{.45\linewidth}
\includegraphics[width=\linewidth]{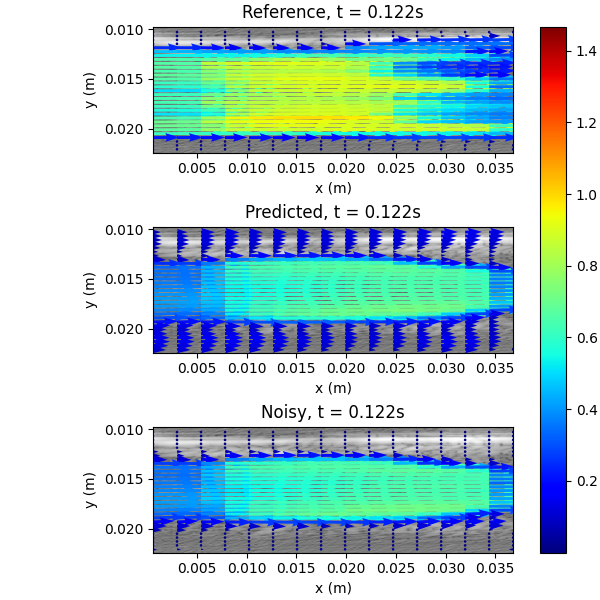}
\caption{}\label{fig:6cm_e_conv_flow2}
\end{subfigure}
\begin{subfigure}[b]{.45\linewidth}
\includegraphics[width=\linewidth]{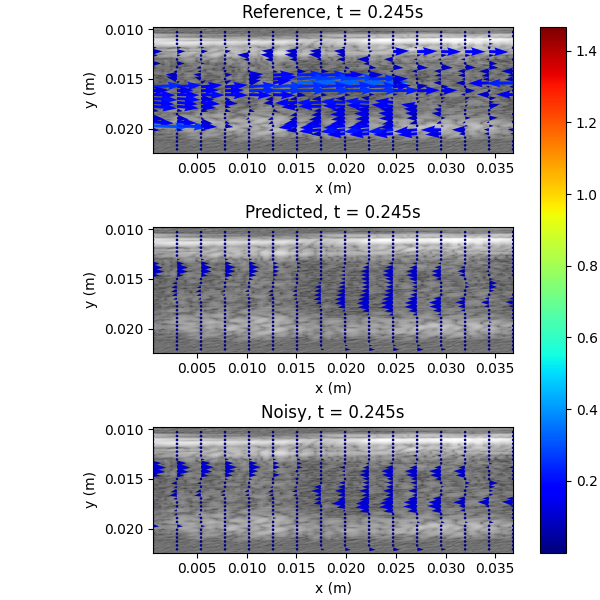}
\caption{}\label{fig:6cm_e_conv_flow3}
\end{subfigure}
\begin{subfigure}[b]{.45\linewidth}
\includegraphics[width=\linewidth]{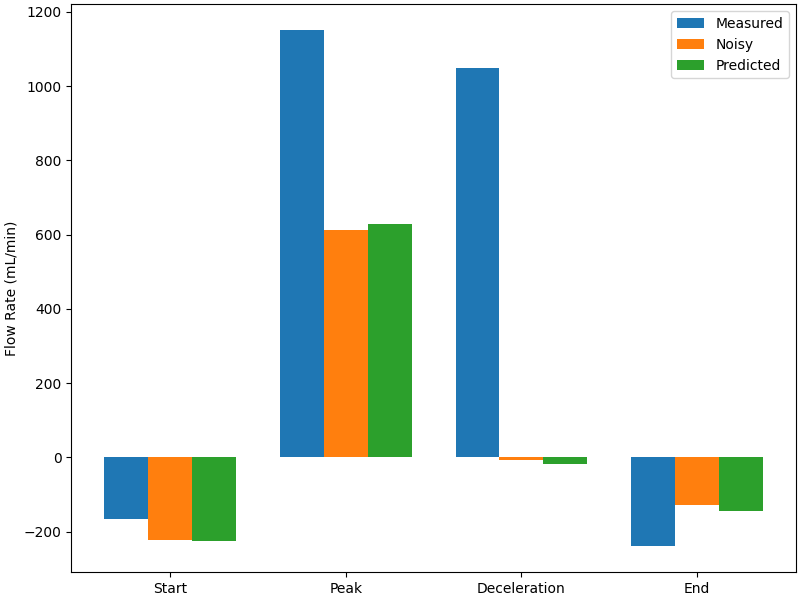}
\caption{}\label{fig:6cm_e_conv_flowrate}
\end{subfigure}
\caption{Qualitative results from applying the RFF neural field to the scan of phantom E with $m=128$ at a depth of 6cm. The flow fields of the reference at 2cm depth (top), prediction from the neural field model (middle), and the noisy dataset (bottom) are shown at times (a) $t=0.020$s, (b) $t=0.122$s, and (c) $t=0.245$s. (d) compares the calculated flow rate with the measured flow rate. Though there was a large difference between the measured and predicted flow rate, the predicted flow field was in agreement with the reference.}
\label{fig:6cm_e_conv}
\Description[Results of Random Fourier Features applied to scan of phantom E at depth of 6 centimetres with low framerate.]{Flow field reconstruction resembles noisy dataset more than the reference flow field, with regions of abnormally high and low velocities retained in the reconstruction but smoothed. Predicted flow rate close to experimental measurements at the start and end of cardiac cycle when flow rate is lowest, but overestimates flow rate at the peak and deceleration stages.}
\end{figure}

For the conventional ultrasound ($m=128$), there was little difference in the flowrate between the RFF model and the Branched model according to Figure \ref{fig:m128}. Though, it appears that the RFF model performs slightly better than the Branched model when comparing the MSE, supporting the conclusion from the experiments with the synthetic data that the RFF model is better for datasets with lower levels of noise. The MSE between the noisy and predicted flow fields improved for all models, but the MSE between the calculated and measured flow rates did not improve at all. Figure \ref{fig:6cm_e_conv} suggests that this could be due to the vast difference between the reference and measured flow rate. The reference flow rate, much like the calculated flow rate, is mostly flat indicating that conventional ultrasound struggles to capture the deceleration accurately. This effectively removes a large amount of flow field data that otherwise could have lead the model to produce a flow rate curve more akin to the measured data. Inpainting these occluded regions could be a significant aid to current ultrasound practices.

\subsection{Inpainting Occluded Flow Fields}

\begin{figure}[h]
\begin{subfigure}[b]{.45\linewidth}
\includegraphics[width=\linewidth]{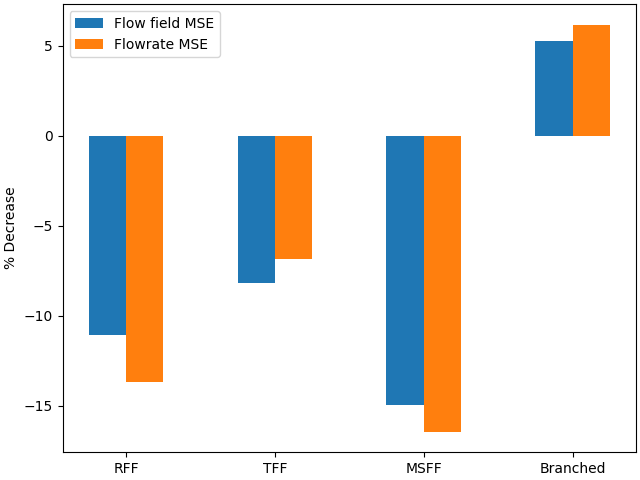}
\caption{}\label{fig:a95_results}
\end{subfigure}
\begin{subfigure}[b]{.45\linewidth}
\includegraphics[width=\linewidth]{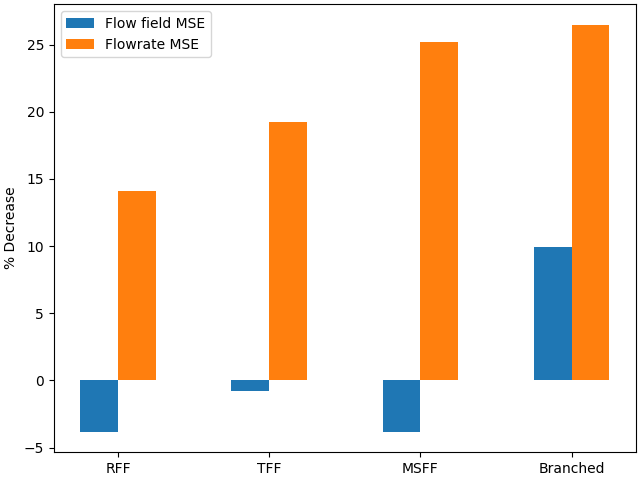}
\caption{}\label{fig:a128_results}
\end{subfigure}
\begin{subfigure}[b]{.45\linewidth}
\includegraphics[width=\linewidth]{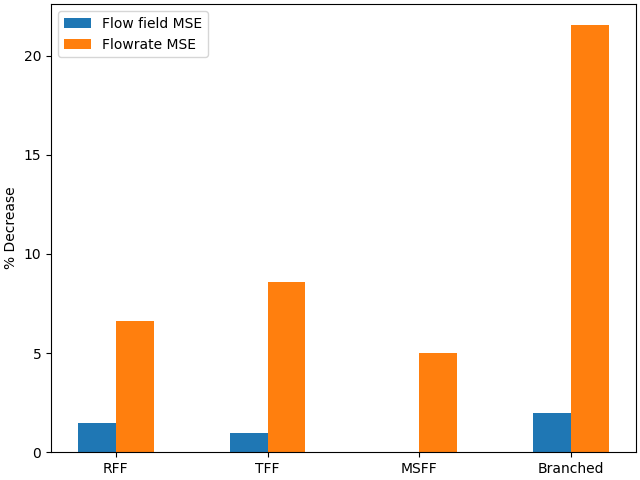}
\caption{}\label{fig:e31_results}
\end{subfigure}
\begin{subfigure}[b]{.45\linewidth}
\includegraphics[width=\linewidth]{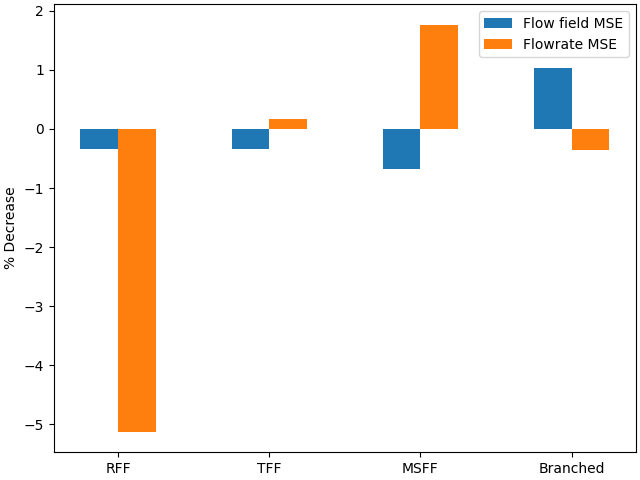}
\caption{}\label{fig:e128_results}
\end{subfigure}
\caption{Relative decrease in flow field and flowrate MSE for the 4 Fourier feature based models when applied to inpainting for (a) Phantom A with $m=95$, (b) Phantom A with $m=128$, (c) Phantom E with $m=31$, and (d) Phantom E with $m=128$. On average, the Branched model achieves the highest relative decrease in MSE when considering both flow field and flowrate.}
\label{fig:inpainting_results}
\Description[Relative performance of 4 Fourier Feature models when applied to inpainting tasks.]{For phantom A with a moderate framerate, only the branched model had a better flow field and flow rate reconstruction than the original noisy dataset. For phantom A with a lower framerate, all models improved the flow rate prediction but only the branched model improved the flow field predictoin too. For phantom E with a high framerate, all models improved flow rate reconstruction and all but the Multi-scale model improved flow field reconstruction, with the best model being the branched model. For phantom E with a low framerate, only the branched model improved the flow field from the noisy data.}
\end{figure}

For inpainting, the focus shifted to using neural field models to improve flow fields with higher $m$ and comparing them with those with lower $m$ for the same flow conditions. Starting with phantom E that was used for denoising ultrasound attenuation, only scans taken at 2cm were considered. Imaging modes with $m\in\{7,31,128\}$ were used in experiments, so the flow fields that were collected when $m\in\{31,128\}$ were reconstructed with the various neural field models and compared with the flow field collected when $m=7$, which was used as a reference. Figures \ref{fig:e31_results} and \ref{fig:e128_results} show the relative decrease in flow field and flowrate MSE for the $m=31$ and $m=128$ scans, respectively. For both phantom E flow fields, the Branched model produced the lowest MSE, albeit a small difference. The figures also show that both the Branched and MSFF models had the lowest MSE when comparing the calculated flow rate with the measured flow rate. 

\begin{figure}[h]
\begin{subfigure}[b]{.45\linewidth}
\includegraphics[width=\linewidth]{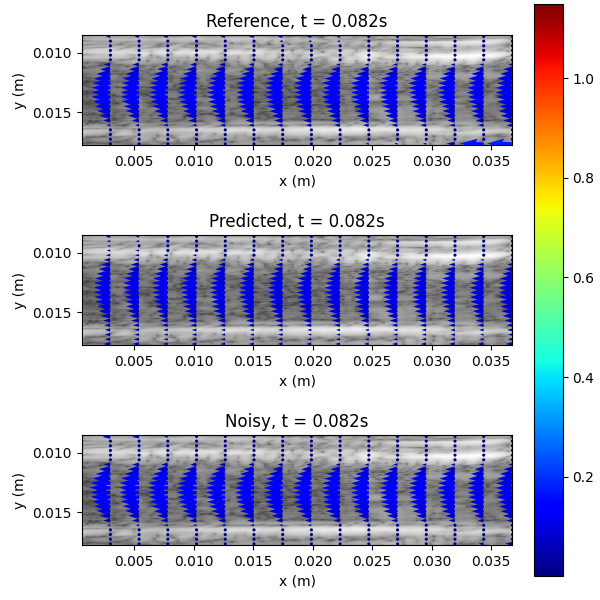}
\caption{}\label{fig:a95_flow1}
\end{subfigure}
\begin{subfigure}[b]{.45\linewidth}
\includegraphics[width=\linewidth]{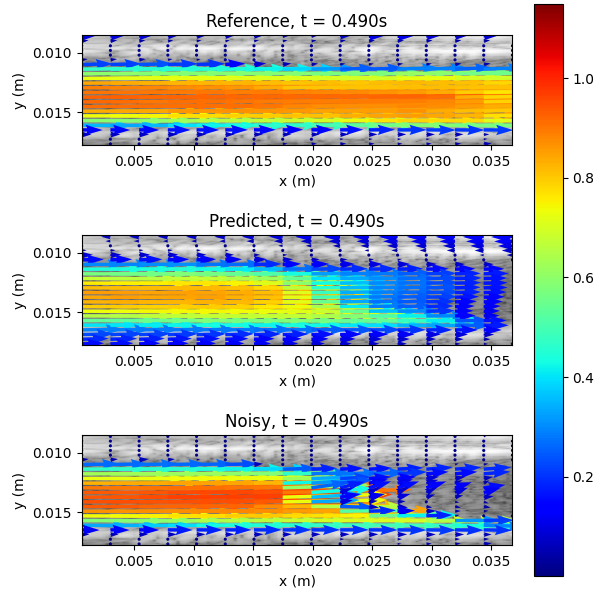}
\caption{}\label{fig:a95_flow2}
\end{subfigure}
\begin{subfigure}[b]{.45\linewidth}
\includegraphics[width=\linewidth]{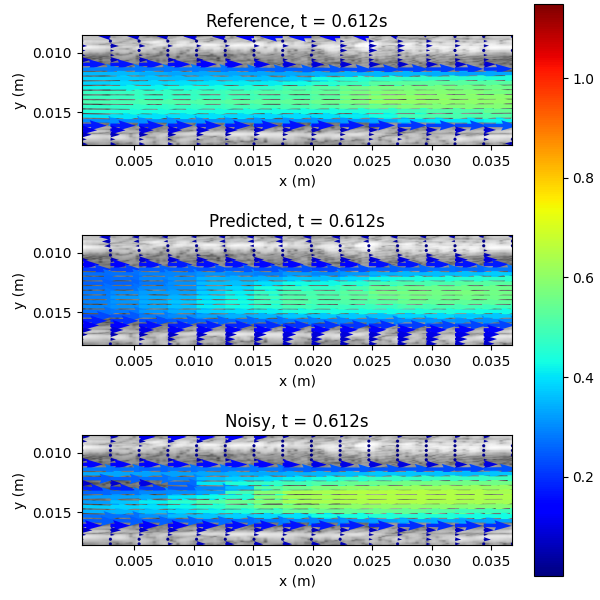}
\caption{}\label{fig:a95_flow3}
\end{subfigure}
\begin{subfigure}[b]{.45\linewidth}
\includegraphics[width=\linewidth]{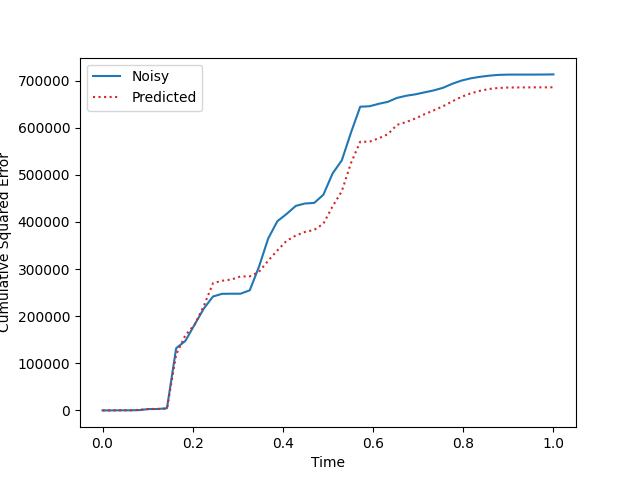}
\caption{}\label{fig:a95_flowrate}
\end{subfigure}
\caption{Qualitative results from applying the branched neural field to the scan of phantom A with $m=95$ at a depth of 2cm. The flow fields of the reference with $m=31$ (top), prediction from the neural field model (middle), and the noisy dataset (bottom) are shown at times (a) $t=0.082$s, (b) $t=0.490$s, and (c) $t=0.592$s. (d) compares the cumulative squared error of the noisy and predicted flowrates. The inpainted flow fields are closer to the reference than the original occluded fields, which also improved the calculated flow rate.}
\label{fig:a95}
\Description[Results of inpainting flow field from phantom A with moderate framerate.]{At low points in the cycle, the prediction is very close to the reference data. As the mean velocity increases, the model inpaints the correct regions of the flow field between the boundary walls but predicts a lower velocity in this region. Inpainted regions are smoother than those in the noisy dataset. Over the cardiac cycle, the predicted flow rate is closer to the experimental measurement than the noisy data.}
\end{figure}

\begin{figure}[h]
\begin{subfigure}[b]{.45\linewidth}
\includegraphics[width=\linewidth]{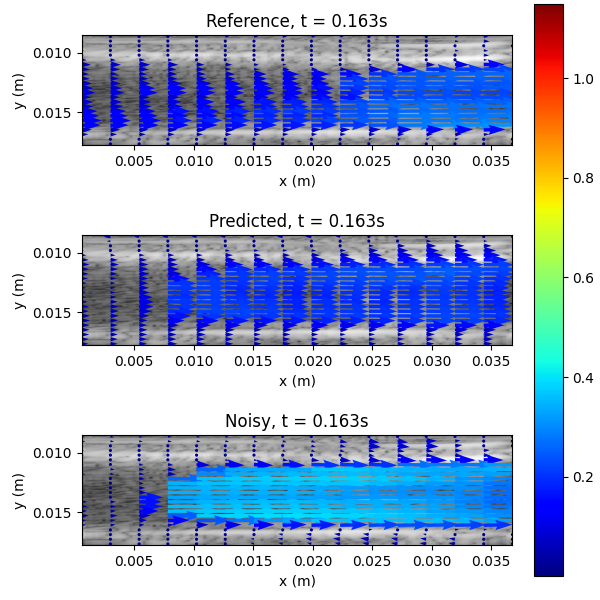}
\caption{}\label{fig:a_conv_flow1}
\end{subfigure}
\begin{subfigure}[b]{.45\linewidth}
\includegraphics[width=\linewidth]{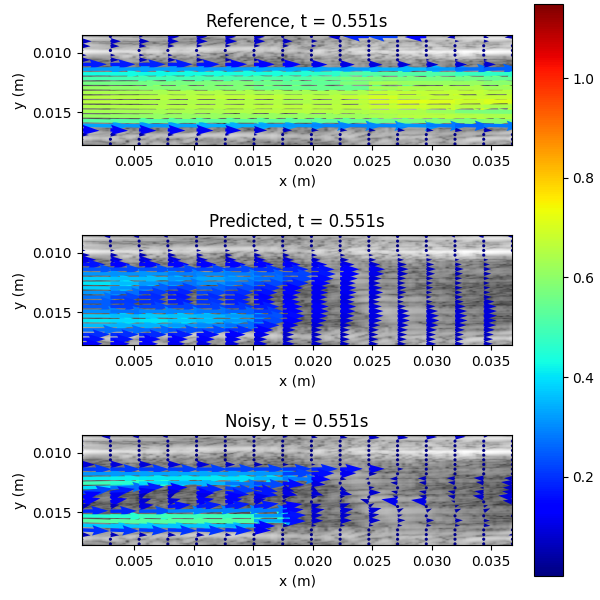}
\caption{}\label{fig:a_conv_flow2}
\end{subfigure}
\begin{subfigure}[b]{.45\linewidth}
\includegraphics[width=\linewidth]{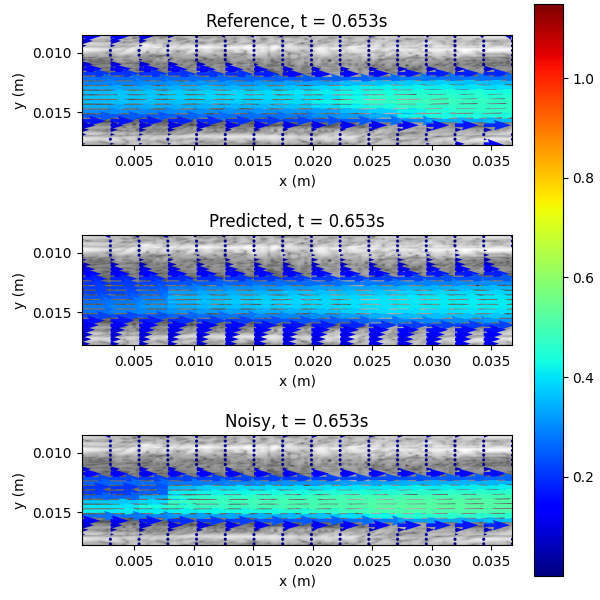}
\caption{}\label{fig:a_conv_flow3}
\end{subfigure}
\begin{subfigure}[b]{.45\linewidth}
\includegraphics[width=\linewidth]{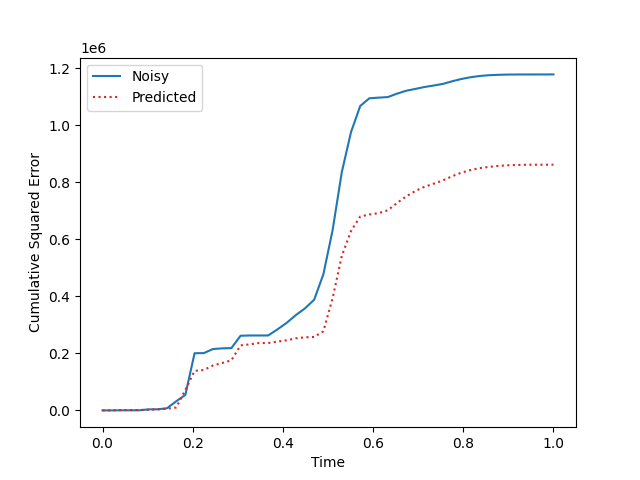}
\caption{}\label{fig:a_conv_flowrate}
\end{subfigure}
\caption{Qualitative results from applying the branched neural field to the scan of phantom A with $m=128$ at a depth of 2cm. The flow fields of the reference with $m=31$ (top), prediction from the neural field model (middle), and the noisy dataset (bottom) are shown at times (a) $t=0.163$s, (b) $t=0.551$s, and (c) $t=0.653$s. (d) compares the cumulative squared error of the noisy and predicted flowrate. Though the inpainting was not as effective as with other datasets, the flow rate has improved during the deceleration phase where conventional ultrasound typically struggles.}
\label{fig:a_conv}
\Description[Results of inpainting flow field from phantom A with low framerate.]{At low points in the cycle, the predicted velocity vectors are close in magnitude to reference data but occluded regions remain. As the mean velocity increases, the occluded regions in the predicted flow field are smaller than they are in the noisy data. There is closest resemblance after the main pulse and the flow has stabilised. Over the cardiac cycle, the predicted flow rate is significantly closer to the experimental measurement than the noisy data.}
\end{figure}

For phantom A, $m=31$ was used as a reference, with $m=95$ and $m=128$ used for denoising. Figures \ref{fig:a95_results} and \ref{fig:a128_results} show that the branched model produced the lowest MSE for flow field reconstruction and flow rate verification for scans with $m=95$ and $m=128$, respectively. Figure \ref{fig:a95} shows the flow field and the flow rate curve for phantom A with $m=95$. The predicted flow field is in general agreement with the reference and the effect of inpainting is clearly seen in Figures \ref{fig:a95_flow2} and \ref{fig:a95_flow3}, where gaps in the noisy flow field have been filled to form a closer resemblance to the reference. Although Figure \ref{fig:a95_flowrate} shows how the noisy and predicted cumulative flow rate errors are very similar, deviation occurs at the peak of the cycle and during the deceleration phase, indicating improvement with the addition of machine learning. Figure \ref{fig:a_conv} shows how much worse the occlusion problem is when $m$ is increased to 128 (conventional ultrasound), as the drop-out regions are less smooth. This caused the model to struggle with reconstruction and inpainting more than with previous datasets, as shown in Figure \ref{fig:a95_flow1}. Despite this, the inpainting model has significantly improved the velocity measurements, as demonstrated by the cumulative flow rate errors in Figure \ref{fig:a_conv_flowrate}. In particular, the deviation between the flowrate of the noisy dataset and that of the predicted flow field coincides with the deceleration phase. This is a part of the cycle that has the highest chance of ultrasound drop-out, as suggested by the flow rate curves from the other scans. To summarise, the MSFF and Branched models performed significantly better than other models, implying that the multi-scale architecture offers an advantage over the other variations of neural fields. 

\section{Conclusion}

Building on the work of \cite{akbariFlowBasedFeatures2021} and \cite{sautoryUnsupervisedDenoisingSuperResolution2024}, which demonstrated the potential of neural fields for reconstructing simulated biomedical flows, and \cite{garziaNeuralFieldsContinuous2024}, which explored their application to real MRI data, this study extends these methods to ultrasound, addressing the unique challenges of flow estimation in this imaging modality. First, the Random Fourier Features (RFF) based neural field architecture proposed by Tancik et al. was extended to trainable and multi-scale alternatives and used to reconstruct arterial flow fields from ultrasound. Preliminary experiments with synthetic data showed promise for the TFF model, suggesting that the claim by Tancik et al. that trainable Fourier Features would offer no advantage could be further explored. It was found, however, that the conclusions from the synthetic data were not entirely transferrable to real data due to different types of noise; notably, occlusions in the flow field. By introducing a branched neural field, that combined the multi-scale model with a vanilla neural field, we consistently achieve the lowest mean squared error across multiple verification metrics (both against a reference flow field and ground truth flow rate measurements) demonstrating robust multi-modal validation. Unlike previous inpainting approaches that rely on a ground truth image for supervision, our models were trained against noisy data only, without clean labels, using physics-based regularisers, marking an important early step in enabling few-shot learning for real-world medical imaging tasks. Furthermore, the successful extension from synthetic to real ultrasound-based flow data represents a critical milestone for translating physics-based AI into real-world applications. Given the limited exploration of physics-informed neural fields to date, this work lays important groundwork for future research at the intersection of physical modeling, self-supervised learning, and data-efficient AI in medicine.

\begin{acks}
VP acknowledges the UKRI Centre for Doctoral Training in Accountable, Responsible, and Transparent AI of the University of Bath (EP/S023437/1).
\end{acks}

\bibliographystyle{ACM-Reference-Format}
\bibliography{neur_bib}

\end{document}